\documentclass[journal]{IEEEtran}
\usepackage[utf8]{inputenc} %
\usepackage[T1]{fontenc}    %
\usepackage{hyperref}       %
\usepackage{url}            %
\usepackage{booktabs}       %
\usepackage{amsfonts}       %
\usepackage{nicefrac}       %
\usepackage{microtype}      %
\usepackage{algorithm}
\usepackage{algorithmic}
\usepackage{graphicx}
\usepackage{makecell}
\usepackage{multirow}
\usepackage{multicol}
\usepackage{bm}
\usepackage{booktabs} %
\usepackage{pdfpages}
\usepackage{overpic}
\usepackage{pict2e}
\usepackage{xcolor}
\usepackage{pifont}
\definecolor{mygray}{gray}{0.45}
\usepackage{textcomp}
\usepackage{times}
\usepackage{graphicx}
\usepackage{amsmath}
\usepackage{amssymb}
\usepackage{pifont}
\usepackage{float}
\usepackage{makecell}
\usepackage{amsmath,amssymb}
\usepackage{mathrsfs}

\RequirePackage{silence}
\graphicspath{{figs/}}

\ifdefined \GramaCheck
\newcommand{\CheckRmv}[1]{}
\newcommand{\figref}[1]{Figure 1}%
\newcommand{\tabref}[1]{Table 1}%
\newcommand{\secref}[1]{Section 1}
\renewcommand{\eqref}[1]{Equation 1}
\else
\newcommand{\CheckRmv}[1]{#1}
\newcommand{\figref}[1]{Fig.~\ref{#1}}%
\newcommand{\tabref}[1]{Tab.~\ref{#1}}%
\newcommand{\secref}[1]{Sec.~\ref{#1}}
\renewcommand{\eqref}[1]{Eqn.~(\ref{#1})}
\fi

\def\ie{\emph{i.e.,~}}
\def\eg{\emph{e.g.,~}}
\def\etc{\emph{etc}}
\def\etal{{~\textit{et al.}~}}
\def\PyrNetName{Illumination-aware Network}

\newcommand{\norm}[1]{\left\lVert#1\right\rVert}
\begin{document}
\title{Designing An Illumination-Aware Network for Deep Image Relighting}
\author{
    Zuo-Liang~Zhu*,
    Zhen~Li*,
    Rui-Xun~Zhang,
    Chun-Le~Guo$^\dag$,
    and~Ming-Ming~Cheng
    \IEEEcompsocitemizethanks{
        \IEEEcompsocthanksitem 
        This work is funded by the 
        National Key Research and Development Program of China (NO. 2018AAA0100400),
        NSFC (NO. 61922046),
        The Fundamental Research Funds for the Central Universities, Nankai University,
        China Postdoctoral Science Foundation (NO.2021M701780).
        \IEEEcompsocthanksitem 
		* denotes equal contribution. 
		\IEEEcompsocthanksitem Z.L.~Zhu, Z.~Li, C.L.~Guo, and M.M.~Cheng
		are with TMCC, CS, Nankai University. 
        C.L.~Guo is the corresponding author (guochunle@nankai.edu.cn).
        \IEEEcompsocthanksitem R.X.~Zhang is with School of Mathematical Sciences, Peking University.
    }
	}

\markboth{IEEE TRANSACTIONS ON IMAGE PROCESSING}%
{Shell \MakeLowercase{\textit{\etal}}: Bare Demo of IEEEtran.cls for IEEE Journals}

\maketitle

\begin{abstract}
    Lighting is a determining factor in photography that affects the style, 
    expression of emotion, and even quality of images.
    Creating or finding satisfying lighting conditions, in reality, is laborious 
    and time-consuming, so it is of great value to develop a technology to manipulate 
    illumination in an image as post-processing. 
    Although previous works have explored techniques based on the physical viewpoint 
    for relighting images, 
    extensive supervisions and prior knowledge are necessary to generate reasonable 
    images, restricting the generalization ability of these works.
    In contrast, we take the viewpoint of image-to-image translation  
    and implicitly merge ideas of the conventional physical viewpoint.
    In this paper, we present an Illumination-Aware Network (IAN) 
    which follows the guidance from hierarchical sampling 
    to progressively relight a scene from a single image with high efficiency.
    In addition, an \textbf{I}llumination-\textbf{A}ware \textbf{R}esidual \textbf{B}lock (IARB) 
    is designed to approximate the physical rendering process and to extract precise descriptors of light sources
    for further manipulations.
    We also introduce a depth-guided geometry encoder for 
    acquiring valuable geometry- and structure-related representations 
    once the depth information is available. 
    Experimental results show that our proposed method produces better quantitative
    and qualitative relighting results than previous state-of-the-art methods.
    The code and models are publicly available on \url{https://github.com/NK-CS-ZZL/IAN}.
\end{abstract}


\IEEEpeerreviewmaketitle

\section{Introduction}
\label{sec:introduction}
\IEEEPARstart{R}{elighting}, which aims to change the illumination settings of an image
 under given lighting conditions, has recently attracted widespread interests
~\cite{sen2005dual, li2018learning, xu2018deep, zhou2019deep, sun2019single,
 nestmeyer2020learning, 2020AIM, 7792614, 9725240, 9785513, 7983410}.
Its high practical value promotes its applications across a variety of fields, 
including mobile imaging, augmented virtual reality, post-processing image editing and 
e-commerce products visualization.

Thanks to the booming of deep learning~\cite{lecun2015deep, simonyan2014very, he2016deep}, 
deep relighting methods
~\cite{sengupta2019neural, li2018learning, xu2018deep, Philip2019Multi, 
nestmeyer2020learning, wang2020single, qiu2020towards, yazdani2021physically, 
xu2018deep, zhou2019deep, sun2019single} 
have significantly accelerated the development in the field of relighting.
With the assistance of powerful representation ability of deep neural networks,
it becomes possible for these methods to relight scenes under more ambiguous 
inputs and complicated environments. 
Most of previous methods
~\cite{xu2018deep, Philip2019Multi, nestmeyer2020learning, wang2020single, qiu2020towards, yazdani2021physically} 
inherit the overall framework of conventional methods and some modified NeRFs
~\cite{srinivasan2020nerv, niemeyer2021giraffe, schwarz2020graf} can also be utilized for relighting. 
However, these methods suffer from huge data requirements to fit a single scene and weak generalization ability. 
So a new framework which can ease constraints on data and 
be utilized in generalized scenarios has been expected. 

Recently, Murmann\etal proposed an indoor scene multi-illumination 
dataset~\cite{murmann2019dataset} which can be used for real scene relighting.
They set up a new point of view in the field of relighting, 
which regards image relighting as an image-to-image translation task.
After Helou\etal proposed a novel outdoor scene synthetic dataset 
(\ie~VIDIT~\cite{helou2020vidit}) and organized the image relighting competitions 
in AIM 2020~\cite{2020AIM} and NTIRE 2021~\cite{helou2021ntire} based on it, 
this viewpoint attracts more attentions.
The viewpoint considers the unavailability of accurate lighting information 
in real-world applications, so only image pairs or triplets with depth information 
are provided for training, lacking specific illumination information. 
Since relighting is highly ill-posed when accurate illumination properties 
are unknown~\cite{ramamoorthi2001signal, aldrian2012inverse}, 
this task therefore becomes more challenging.
For example, Ramamoorthi and Hanrahan~\cite{ramamoorthi2001signal} demonstrate that the difference 
between low-frequency texture and lighting effects is hard to distinguish for most situations.  

To tackle relighting on such ill-posed condition, we hope to get inspiration from rendering frameworks.
However, instead of previous physics-based networks 
which directly estimate rendering-related parameters with supervisions by laboriously designed losses.
We intend to figure out specific network and modules corresponding with rendering process 
and design a network intrinsically suitable to relighting 
without supervisions for intermediate parameters.
We mainly establish links between our proposed network and typical ideas of conventional rendering 
in following aspects: 
\textbf{1)} Hierarchical sampling strategy~\cite{levoy1990efficient}, which 
has shown its ability and efficiency in voxel rendering. 
\textbf{2)} Spherical harmonic lighting~\cite{green2003spherical},
which parameterizes light source with bases of different frequencies. 
\textbf{3)} Physics-based rendering under the spherical harmonic 
lighting~\cite{green2003spherical, imageworks2010physically},
which can be modeled without integral and needs only multiplication.

According to above discussions, 
we propose an \textbf{I}llumination-\textbf{A}ware \textbf{N}etwork (\textbf{IAN}) 
for accurate deep image relighting in this paper.
Specifically, as a \textbf{simulation of hierarchical sampling}, 
a pyramid-like architecture is deployed for progressively 
changing the lighting condition in an image from coarse to fine.
In addition, 
inspired by the idea of physics-based rendering model, 
we elaborately design an illumination-aware residual block in a two-branch structure. 
With the guidance from \textbf{spherical harmonic lighting} 
which decouples light into components of different frequencies,
we utilize convolutions with diverse dilation rates to obtain samples under different frequencies 
and design a statistical-coupled attention branch as illumination descriptor extractor, 
which models light by diverse statistics.
Another branch preserves local geometry- and reflectance-related information.
Finally, the multiplication of the illumination descriptor 
and the geometry-related information
implicitly serves as an approximation of \textbf{rendering 
under spherical harmonic lighting assumption}.
Besides, considering that depth information is available in many applications, 
we also introduce a depth-guided geometry encoder which takes the 
depth map, surface normal, and positional encoding as the inputs, 
aiming to extract multi-scale geometry and structure information 
to assist the relighting task.

We evaluate the proposed network on the VIDIT~\cite{helou2020vidit} dataset
with the absence and the presence of depth information,
which correspond to the settings of AIM 2020~\cite{2020AIM} 
and NTIRE 2021~\cite{helou2021ntire} challenges, respectively.
Our proposed method outperforms all comparison methods, including the champion solutions of 
AIM 2020~\cite{2020AIM} and NTIRE 2021~\cite{helou2021ntire} both quantitatively and quantitatively.
Besides, we also perform evaluations on the Adobe Multi-Illumination dataset 
which contains real indoor scene, 
and our results still obtain the best performance in comparison with other methods.
Then we apply our method on a portrait relighting dataset (\ie DPR dataset~\cite{zhou2019deep}) 
and our method surpasses previous methods by a large margin, 
demonstrating the superiority and robustness of our proposed method.

In summary, our contribution is three-fold:
\begin{itemize}
\item We design an illumination-aware network (IAN) 
which implicitly inherits the idea of physics-based rendering to perform image relighting.
Through extensive experiments, we show that our proposed method achieves 
better performance than other methods while maintaining promising computational efficiency.
\item We propose an illumination-aware residual block (IARB) which implicitly conducts rendering process 
and is suitable for relighting task.
\item We introduce a depth-guided geometry encoder to fully extract the geometry-
and structure-related features from additional information (\eg depth, normal, and linear positional encoding). %
These features assist the network to obtain favorable relighting results.
\end{itemize}

\section{Relative Works}
Numerous image relighting methods have been proposed in the literature.
Based on the usage of convolutional neural network, 
we divide these methods into two groups: conventional physics-based method 
and deep network based methods.

\subsection{Conventional physics-based methods}
Conventional physics-based methods focus on building explicit assumptions and models 
to approximate effects of illumination in reality efficiently.
These assumptions~\cite{basri2003lambertian} greatly reduce dimensions of light transport function 
into low-dimensional subspace~\cite{basri2003lambertian, belhumeur1998set, ramamoorthi2001relationship} 
in order to ease the difficulties in calculation.

Though dimensions of light transport function are highly reduced, 
fitting a reasonable function still needs hundreds of images of a scene 
by brute-force searching~\cite{debevec2000acquiring, sen2005dual}.
Subsequently, to reduce the number of required samples to perform relighting, 
some works~\cite{wang2009kernel, reddy2012frequency} take advantages of 
the local coherence of light transport matrix in lower dimensions~\cite{malzbender2001polynomial} 
and others~\cite{karsch2011rendering} involve in human interactions.
Decomposing rendering-related factors from given images of a scene
~\cite{karsch2011rendering, duchene2015multi, zhang2016emptying} 
is a widely used strategy in relighting, which is known as inverse renderer. 
Geometry, material, and illumination are usually predicted separately at first.
By directly controlling these explicit factors, these methods can re-render given scene and 
obtain relighting results of good quality.

However, these methods require a complex calibration process, huge computational costs, storage resources,
and even specialized hardware (\eg panorama camera and drone).
Rendering-related factors (\eg geometry, surface reflectance, and environmental illumination) are either 
estimated by complicated system or measured by specific equipment. 
The former leads to the accumulation of errors in the whole process 
and the latter limits the general applications of these methods.
Besides, numerous input images with strict constraints are needed to fit a model for a single scene.
In contrast, our method is data-driven, and the training data is easy to acquire and access.
Once the model is trained, only a single image is needed to perform relighting 
and the pretrained model can be generalized into diverse scenes.

\subsection{Deep network based methods}
Recently, deep neural networks~\cite{krizhevsky2012imagenet,he2016deep} have shown 
their potentials on illumination-related
manipulation~\cite{ding2019argan, nagano2019deep,zhang2020portrait,zhang2020copy}, 
which promotes the development of deep network based methods for relighting task. 
There are mainly three viewpoints to design relighting networks, 
namely physics-based viewpoint, neural radiance field viewpoint, 
and image-to-image translation viewpoint.

\subsubsection{\textbf{Physics-based viewpoint}}
    Physics based neural network derives from conventional physics-based methods and 
    these methods replace parts of original system with neural networks.
    Owing to the strong representation ability of neural network, 
    Ren \textit{et al.}~\cite{ren2015image} and Xu \textit{et al.}~\cite{xu2018deep} 
    simplify the estimation process of light transport function with sparse samples.
    Inspired by the idea of decomposition, some methods~\cite{Philip2019Multi, 
    nestmeyer2020learning, wang2020single, qiu2020towards, yu2020self} 
    employed different networks with the guidance of corresponding losses to factor an image 
    into multiple components, including albedo, normal, shading, \etc.
    To avoid accumulation of errors in decomposition and re-rendering procedure, 
    some concurrent methods only insert the lighting priors 
    (\ie spherical harmonics~\cite{xu2018deep, zhou2019deep} and environment map~\cite{sun2019single}) 
    into the network for directly obtaining the desired relighting results.
    These methods also inherit disadvantages of conventional physics-based methods. 
    For instance, multiple calibrated images and lighting priors are expensive 
    and laborious to acquire and usually absent in real-world applications. 
    Though the accumulation of errors is eased by robustness of neural network, 
    it is still a problem pending to be further solved.
    
\subsubsection{\textbf{Neural radiance field viewpoint}}
    Neural radiance fields directly construct continuous representations for scenes, 
    parameterized as a basic MLP network. 
    Since Mildenhall \textit{et al.} proposed NeRF~\cite{mildenhall2020nerf}, numerous works 
    have attempted to optimize it. 
    Some works~\cite{srinivasan2020nerv, martin2021nerf} improve the abilities 
    or extend the application scenarios of vanilla NeRF~\cite{mildenhall2020nerf}.
    NeRV~\cite{srinivasan2020nerv} enhances the ability for recovering relighting 3D scene representations 
    and NeRF in the wild~\cite{martin2021nerf} succeeds in modeling ubiquitous, 
    real-world phenomena in uncontrolled images.
    Attributed to the representation ability of GAN, 
    GIRAFFE~\cite{niemeyer2021giraffe} and GARF~\cite{schwarz2020graf} make scenes editable.
    These successive works enable to manipulate light condition in a scene, which can be used 
    in relighting.
    The major disadvantage of these methods is their limited ability in generalization. 
    One pretrained model can only take effect on a single scene. 
    Besides, their inputs contain 3D camera pose and position information, 
    which is inaccessible in some scenarios.
    In the contrary, our method needs no position information of real-world and can take effect on 
    various scenes. 
    Our method also benefits from few parameters and low computational cost.

\subsubsection{\textbf{Image-to-image translation viewpoint}}
    Murmann\etal~\cite{murmann2019dataset} attempt to discard explicit graphics prior and 
    regarded the image relighting as the image-to-image translation problem.
    Shared with similar settings,
    Helou\etal proposed a virtual image dataset~\cite{helou2020vidit} for illumination transfer 
    and held image relighting competitions (\ie AIM 2020~\cite{2020AIM} and NTIRE 2021~\cite{helou2021ntire}). 
    The competitions mainly include two tracks named ono-to-one relighting and any-to-any relighting. 
    These proposed datasets and competitions motivate researchers to think relighting task from a brand-new viewpoint, 
    namely image-to-image translation viewpoint.
    Some works involve and adjust existing modules or networks 
    which have shown their representation abilities in other fields. 
    Puthessery\etal, winners of AIM 2020, proposed WDRN~\cite{puthussery2020wdrn} 
    which employs the wavelet transformation for efficient multi-scale representations.
    Paul\etal~\cite{gafton20202d} applied pix2pix~\cite{isola2017image} to their framework 
    and utilize adversarial learning to further improve the quality of the generated images.
    Yang\etal~\cite{yang2021multi} took the corresponding depth map into consideration and 
    designed a depth-guided relighting network based on an RGB-D saliency detection method~\cite{pang2020hierarchical}.
    This type of methods is easier to train than ones from another two viewpoints, 
    for they relieve the constraints exerted on input. 
    These methods consider relighting of general cases, 
    which makes general applications possible.
    However, existing methods underperform above-mentioned two viewpoints and 
    how to fully explore their ability remains to be solved. 
    Our work follows image-to-image translation viewpoint and extends its boundary of performance. 
    Besides, we blend physics-based ideas implicitly and design modules intrinsically 
    suitable to relighting task.

\begin{figure*}[ht]
    \centering
    \includegraphics[width=0.95\textwidth]{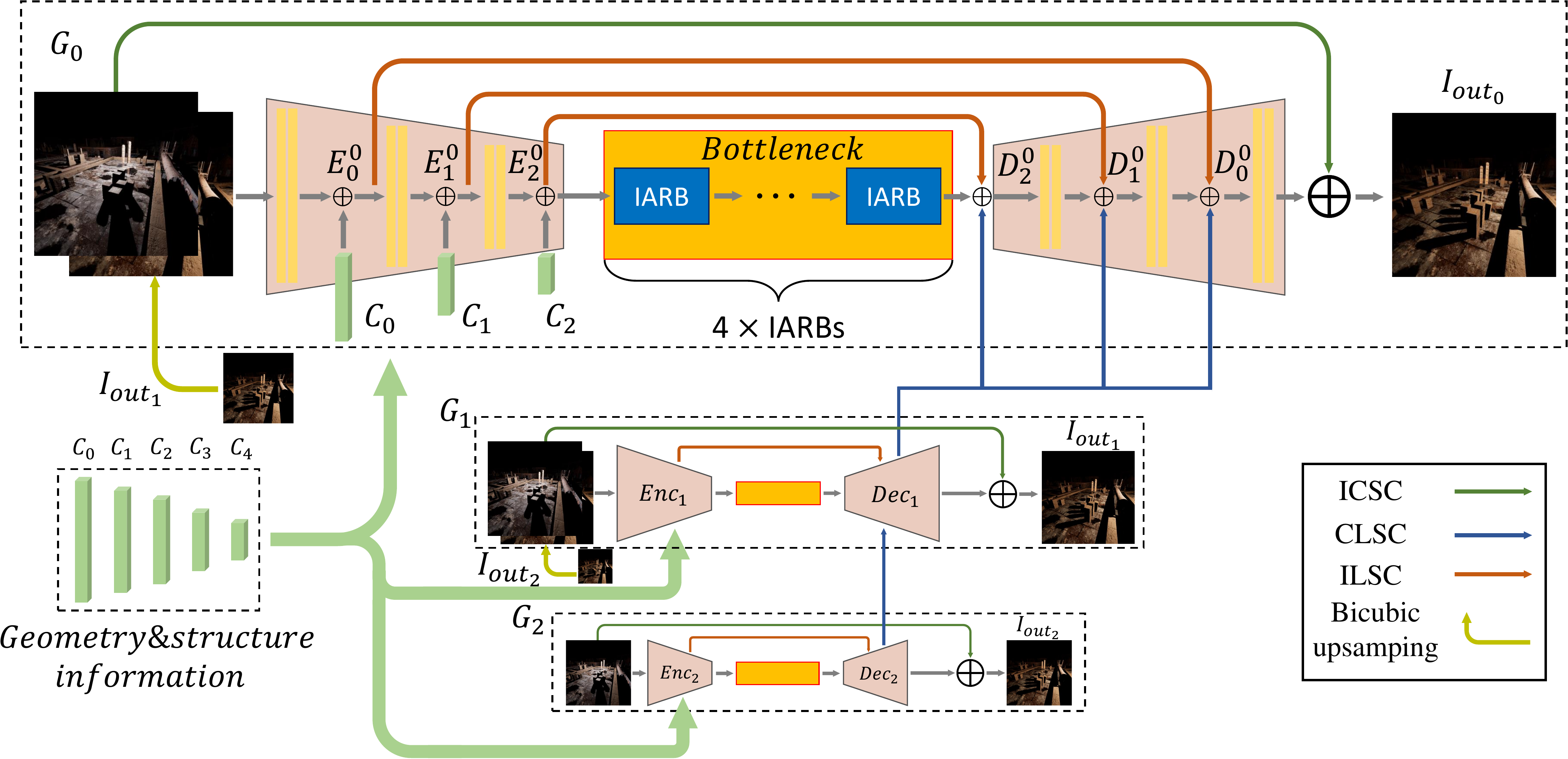}
    \caption{
    Overview of the proposed illumination-aware network (IAN). 
    The proposed network is designed as a pyramid-like structure and processes images in a coarse-to-fine manner. 
    Geometry- and structure-related information is provided from depth-guided geometry encoder 
    which will be demonstrated in \secref{ssec:aux_enc}.
    In this figure, ICSC, ILSC and CLSC denote image content, intra-level, 
    and cross-level skip connections, respectively. IARB represents the proposed illumination-aware residual block
    which will be shown in \secref{ssec:res}.
    The output of the $l$-th level is up-sampled by bicubic interpolation and is concatenated with unprocessed image 
    as input for the $(l-1)$-th level.
    }
    \label{fig:overview}
    \vspace{-5mm}
\end{figure*} 

\section{Approach}

We define that $I_{in}$ represents the input image 
under a pre-defined illumination condition. 
$I_{gt}$ represents the ground truth under 
the desirable illumination condition. 
Single image relighting aims to translate the input image $I_{in}$
to another image $I_{out}$ 
whose illumination condition is similar with $I_{gt}$ through a relighting network $G$.
Considering relighting task commonly does not take substantial 
environmental changes into account, 
so above images are from the same scene if without particular explanation.

In terms of task definition, we partially follow AIM 2020~\cite{2020AIM} and 
NTIRE 2021~\cite{helou2021ntire} competitions, which divides relighting task
into 2 cases, namely one-to-one relighting and any-to-any relighting.
We firstly focus on the former case in the competitions~\cite{2020AIM, helou2021ntire}
where input and target illuminant settings are pre-determined and fixed for all scenes. 
In this setting, the output image is formulated as $I_{out} = G(I_{in}, g_{opt})$ 
where $I_{in}$ is the input and $g_{opt}$ is some optional guidance.
Besides, the any-to-any relighting setting in the competitions~\cite{2020AIM, helou2021ntire} 
has only 40 pre-defined illumination condition and cannot fully validate the ability of our model in 
arbitrary illumination relighting.
So we consider a more generalized setting that contains a continuous space of illumination 
and extend our method to this condition.
An additional target light prior $l_p$ is needed,
 and the output is formulated as $I_{out} = G(I_{in}, g_{opt}, l_{p})$ 

In this section, we present an illumination-aware network (IAN)
which enables high-resolution relighting image generation of arbitrary scenes.
Specifically, we propose a pyramid-like network architecture (see \secref{ssec:pyr}) along with residual learning strategy.
This network architecture progressively manipulate effects of light 
in order to generate relighting images with fine-grained details and global consistency.
Besides, an illumination-aware residual block (IARB) (see \secref{ssec:res}) is proposed to parameterize 
attributes of light source and to leverage the extracted illumination descriptor 
for an implicit rendering process.
To further utilize depth information, which can be estimated from RGB images or 
captured by advanced sensors, we propose a depth-guided geometry encoder (DGGE) (see \secref{ssec:aux_enc}) shared among levels.


\subsection{\PyrNetName}
\label{ssec:pyr}

As the main body of our network (see Fig.~\ref{fig:overview}), 
a pyramid-like network architecture~\cite{zhang2019deep, das2020fast} 
utilizes multi-scale information from input images 
and conducts image relighting in a coarse-to-fine manner. 
We will introduce this architecture in this section.

For relighting task, 
owing to the diverse effects of light which entangle with scene attributes, 
complex information from low-level features (\eg texture and edge) to high-level features 
(\eg class of object) is needed and accounts for the final relighting results. 
To capture information in various semantic levels, 
we utilize U-Net~\cite{ronneberger2015u} like structure 
which is proven to be effective in numerous previous 
works~\cite{zhou2019deep, sun2019single, puthussery2020wdrn}.
However, a single U-Net is insufficient to tackle with relighting task.
Firstly, 
due to the highly ill-posed property of relighting 
and diverse effects of light among scales,
achieving high-quality relighting results by a single step is 
too complicated for a single U-Net.
Besides, since the resolution of an input image is high 
and a single-level network has a limited receptive field, 
such network only concentrates on local features and ignores global cues.
Since local features are easily affected by the material and color of the object, 
it is hard to extract intrinsic attributes of light from them. 
Consequently, 
we observed that the results from vanilla U-Net~\cite{ronneberger2015u} are 
trapped into local minima
due to the disability of vanilla U-Net in capturing global light information.

Considering above-mentioned difficulties, 
we resort to traditional rendering framework to get inspiration.
In the field of voxel rendering, 
hierarchical sampling strategy~\cite{levoy1990efficient} is designed to tackle similar problems. 
This strategy arranges voxel rendering in a progressive manner 
to fulfill this process effectively and efficiently.
As previous works revealed~\cite{yazdani2021physically, chen2020neural}, 
relighting task can be seen as a re-rendering process. 
So hierarchical sampling strategy intrinsically benefits to relighting task.
Moreover, humans tend to focus on overall low-frequency and tone changes
before they take local structures into account, 
which refers to a global-to-local architecture.

Motivated by above rendering strategy and human preference, 
we further extend U-Net~\cite{ronneberger2015u} to a pyramid-like architecture.
Compared with the previous one, 
the receptive field of this network is tremendously enlarged 
and is sufficient to capture global information.
This design eases the difficulties of task assigned to each pyramid level 
and promotes the quality of final relighting results.

This pyramid-like architecture has $3$ levels in total noted as $G_0$, $G_1$, $G_2$ 
from bottom to top. 
The full resolution input image and the $2\times/4\times$ bicubic down-sample ones are denoted as 
$I_{in}$, $I_{in}^{\downarrow 2}$ and $I_{in}^{\downarrow 4}$, respectively.
The outputs of $G_0$, $G_1$, $G_2$ are $I_{out_{0}}$, $I_{out_{1}}$ and $I_{out_{2}}$.
$G_0$ takes $I_{in}$ and $I_{out_{1}}^{\uparrow 2}$ which is 
$2\times$ bicubic up-sampled output of $G_1$ as input. 
$G_1$ takes $I_{in}^{\downarrow 2}$ and $I_{out_{2}}^{\uparrow 2}$ which is
$2\times$ bicubic up-sampled output of $G_2$ as input. 
$G_{2}$ only takes $I_{in}^{\downarrow 4}$ as input, 
for no previous output is available.
For each level, 
an encoder down-samples features for 2 times and a decoder up-samples them correspondingly, 
which resembles a U-Net~\cite{ronneberger2015u}. 
A bottleneck comprised of 4 illumination-aware residual blocks bridges the encoder and the decoder.
We will detail this residual block in Sec.~\ref{ssec:res}.

To alleviate illumination invariant information loss during encoding, we utilize intra-level skip 
connections (ILSC) as: 
\begin{equation}
        D_i = D_i + E_i,
\end{equation}
where $D_i$ and $E_i$ denote the features from encoder and decoder 
at the same pyramid level, respectively.
$i$ represents the times of down-sampling at current level. 
Since all features involved in the above equation are 
in the same pyramid level, we ignore the superscript 
which presents the number of pyramid level.

As higher levels already modeled global illumination changes, 
we preserve them when modeling local influences of light. 
Besides, every level is similarly assigned to relight image in a residual manner, 
so information is intrinsically shared among levels. 
It is more reasonable to refine features of previous level 
instead of encoding brand-new features in each level. 
To achieve the proposals, we introduce a cross-level skip connection (CLSC).
This skip connection directly feeds information from a smaller scale
to a larger one at the decoder side 
in order to reinforce scale-specific information learning in current resolution.
This strategy also contributes to decoupling light effects on different scales, 
for the upper level only needs to take charge of modeling global information, 
regardless of local details that lower levels take charge of.
The CLSC can be formulated as:
\begin{equation}
    D^l_i = D^l_i + [D^{l+1}_i]_{\uparrow 2},
\end{equation}
where $l$ represents the level in the pyramid-like structure, 
and $[D^{l+1}_i]_{\uparrow 2}$ represents bilinear up-sampled features for 2 times from previous level.

Although the CLSC succeeds in preserving illumination invariant information, detailed textural 
information is hard to fully reconstruct during decoding.
As the domain of inputs and outputs of the network are consistent,  
differing from the vanilla U-Net~\cite{ronneberger2015u} designed for segmentation, 
an image content skip connection (ICSC) is used to directly deliver input images 
to the output side of network for retaining fine-grained textural details.
Besides, the CLSC makes decoders aware of cross-scale information, while encoders are blind 
to it. 
To extract the most beneficial features at the encoder side, 
up-sampled output from previous level is taken as part of input, 
which helps encoders be aware of cross-scale information as well. 
Eventually, entire pipeline is formulated as:
\begin{equation}
    \begin{aligned} 
    I_{out_{l}} &=
    &
    \left\{
    \begin{aligned}   
    &G_{l}(I_{in}^{\downarrow 2^{l}}) + I_{in}^{\downarrow 2^{l}} & l=2 \\
    &G_l(\mathbf{cat}(I_{in}^{\downarrow 2^l}, I_{out_{l+1}}^{\uparrow 2})) + I_{in}^{\downarrow 2^l} & otherwise
    \end{aligned}
    \right.
    \end{aligned}
\end{equation}
where $\mathbf{cat}$ represents a concatenation operator.

\subsection{Illumination-aware residual block}
\label{ssec:res}

\begin{figure}
    \centering
    \includegraphics[width=0.48\textwidth]{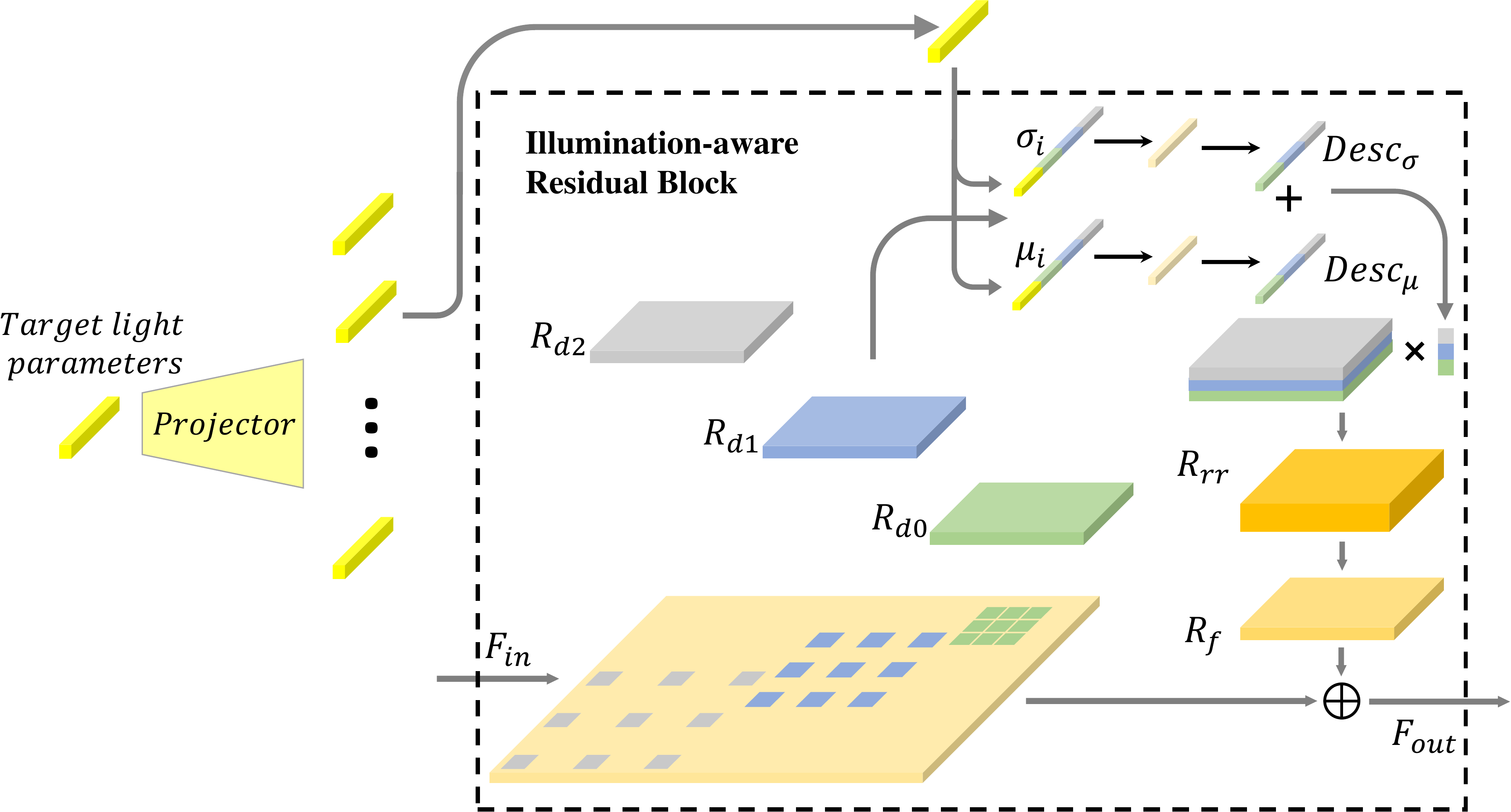}
    \caption{Structure of the proposed illumination-aware residual block (IARB). 
    This delicate block implicitly embeds rendering process and is able to utilize 
    various statistics under diverse sampling frequencies 
    to obtain an accurate light descriptor. 
    Along with a target light projector,
    IARB further enables the manipulation of light condition.}
    \label{fig:res}
    \vspace{-5mm}
\end{figure} 

Physics-based rendering model which has shown its strong ability in computer graphics is defined as:
\begin{equation}
    \label{equ:brdf}
    I(x,\omega_o)=\int_{\Omega}f(x,\omega_i,\omega_o)L(x,w_i)n\cdot\omega_id\omega_i,
\end{equation}
where $f(x,\omega_i,\omega_o)$ denotes Bidirectional Reflectance Distribution Function, 
$L(x,w_i)$ is the radiance that arrives at the point $x$ from the incoming direction $\omega_i$, 
$n$ is the surface normal at current position, and $\omega_o$ is the outgoing direction.
Inspired by this, we attempt to design a module to implicitly embed this process into our network, 
as shown in \figref{fig:res}.
Due to the difficulties in directly calculating integral in \eqref{equ:brdf},
we utilize the spherical harmonic lighting to approximate original light condition. 
We replace the integral and  $L(x,w_i)$ with $c_j \odot Y_j$ 
where $Y_j$ is the $j$-th spherical harmonic basis 
and $c_j$ is its coefficient.
Finally, \eqref{equ:brdf} is rewritten as \eqref{equ:brdf_sim}: 
\begin{equation}
    \label{equ:brdf_sim}
    I(x,\omega_o)= \sum_j \underbrace{[f(x,\omega_i,\omega_o) \odot (n\cdot\omega_i)]}_{surface~attributes} \odot \underbrace{[c_j \odot Y_j]}_{light},
\end{equation}
where $\odot$ represents the element-wise scalar product and 
$\cdot$ is the dot product.

The \eqref{equ:brdf_sim} can be divided into 2 parts. 
The first part $f(x,\omega_i,\omega_o) \odot (n\cdot\omega_i)$ 
is mainly related to the geometry and texture attributes of objects.
The second part $c_j \odot Y_j$ describes the attributes of light. 
Corresponding to the two parts of~\eqref{equ:brdf_sim}, 
we hope to design a module which has two abilities accordingly. 
One is the illumination-aware ability.
We hope this module to extract a credible spatial-invariant illumination descriptor 
which represents a specific group of spherical harmonic coefficients implicitly.
Then this descriptor is projected into a desirable descriptor 
which offsets the effects of old illumination and exerts influences of new one. 
Another is the geometry-maintenance ability.
This module should preserve local surface features 
which contain textures, surface normals, and positional information.

We first describe how to design a module 
that can extract the light descriptor under 
the spherical harmonic lighting assumption, which represents the lights by 
several bases in different domain. 
The coefficients of bases in spherical harmonic lighting are calculated 
as~\eqref{eq:sh}:
\begin{equation}
    c_j = \sum_\omega^\Omega{F_{light}(\omega)\cdot Y_j},
    \label{eq:sh}
\end{equation}
where  
$F_{light}(\omega)$ is the function which represents the global illumination 
and $\Omega$ is the set of frequencies.

Noted that the convolution in the spatial domain is the multiplication 
in frequency domain, we intuitively utilize kernels with diverse dilation 
rates which bring diverse sample rates as different bases in spherical harmonic 
lighting. 
In practice, we utilize dilation rates ranging from $1$ to $3$ and stack several 
modules to simulate dilation rates in a larger range.
The outputs of dilation convolutions are formulated as:
\begin{equation}
    R_{ori} = \{R_{d0}, R_{d1}, R_{d2}\}.
\end{equation}
Besides, the concatenation and the linear projection can be seen as a generalized 
case of summation (The summation can be written as $sum=\mathbf{w^T} f$ 
when $\mathbf{w^T}=\mathbf{1^T}/N$, where $f$ is the concatenated feature and 
$N$ is the dimension of $f$). 
Thus, we replace the summation with the concatenation and the linear projection, 
and we convert the convolution in the spatial domain to the multiplication in the frequency domain 
to build the connection between the proposed illumination-aware residual block (IARB) with~\eqref{eq:sh}. 
The approximation of~\eqref{eq:sh} in our IARB can be formulated as: 
\begin{equation}
    Desc_\mu = \mathbf{Linear}(\mathbf{cat}(f_k \ast f_f)) \xrightarrow{\mathscr{F}}
    \sum_k^{k\in\mathscr{K}}  [w_k \cdot F_f(\omega)]\cdot  F_k(\omega),
    \label{eq:conv_sum}
\end{equation}   
where $F_k(\omega)$ and $F_f(\omega)$ are the representation of 
convolution kernels and features in the frequency domain respectively,  
$f_k$, $f_f$ are their representations in the spatial domain respectively ($f_k \ast f_f \in R_{ori}$),
$w_k$ is the weight in linear projection for the $j$-th kernel,
$\mathscr{K}$ is the set of kernels, $\mathscr{F}$ presents the Fourier transformation 
and $\ast$ is the convolution operation. 
Now we obtain the first item $Desc_\mu$ to describe attributes of light source. 
Besides, we also introduce $Desc_\sigma$ which is the deviation of features
to measure non-linear relations for a better model of lighting. 
Then
considering the invariance of the illumination condition in a scene, 
we conduct a global average pooling before linear projection to obtain a 1D feature 
to diminish influences caused by spatial positions. 
In practice, $Desc_\mu$ and $Desc_\sigma$ are calculated as: 
\begin{equation}
    Desc_{\mu} = F_{\mu}(\mu), \mu=[\mu_1,...\mu_c],
\end{equation}
\begin{equation}
    \label{equ:mean}
    \mu_i = \frac{1}{H\times W}\sum_{j, k}^{H, W}{{R_i(j, k)}},
\end{equation}
\begin{equation}
    Desc_{\sigma} = F_{\sigma}(\sigma), \sigma=[\sigma_1,...,\sigma_c],
\end{equation}
\begin{equation}
    \label{equ:std}
    \sigma_i = \sqrt{\frac{1}{H\times W}\sum_{j, k}^{H, W}{(R_i(j, k)-\mu_i)^2}},
\end{equation}
where $\mu_i$ and $\sigma_i$ are the mean and standard deviation of the $i$-th channel of 
features $R_{ori}$ respectively.

Except for the branch to extract attributes of light, 
another branch is designed to preserve spatial information 
correlated with normals, textures.
Eventually, this module has two components corresponded 
to two above-mentioned abilities, and we obtain two items which 
represent surface attributes and light conditions respectively as:
\begin{equation}
    \frac{Desc_{\mu} + Desc_{\sigma}}{2} \sim [c_j \odot Y_j], 
\end{equation} 
\begin{equation}
    R_{ori} \sim [f(x,\omega_i,\omega_o) \odot (n\cdot\omega_i)].
\end{equation} 
So the rendering process in~\eqref{equ:brdf_sim} 
express as \eqref{equ:rere} in network and $R_{rr}$ is the re-rendered feature:
\begin{equation}
    R_{rr} = R_{ori} \odot \frac{Desc_{\mu} + Desc_{\sigma}}{2}.
    \label{equ:rere}
\end{equation}
For matching the original feature space, 
we use a $3\times 3$ convolutional layer 
to compress the re-rendered feature $R_{rr}$.
Eventually, due to the orthogonality of spherical harmonic lighting, 
this module can modify a subset of lighting components without influencing others. 
We thus can simply add the re-rendered feature to the original one, 
and the output feature $F_{out}$ is calculated as:
\begin{equation}
    F_{out} = C_f(R_{rr}) + F_{in}.
\end{equation}
To fully model light conditions under a spherical harmonic setting, 
we utilize multiple modules to re-render different lighting components  
as the summation process in~\eqref{equ:brdf_sim}.

In order to further enable manipulation of light condition, 
we design a target light projector which receives parameterized light as input
and produces its descriptors as illumination guidance for IARBs. 
The modified IARB for manipulation of light condition calculates the illumination 
descriptor as
\begin{equation}
    Desc_{\mu/\sigma} = F_{\mu/\sigma}(\mathbf{cat}(\mu/\sigma, l_p)),
\end{equation}
where $l_p$ is the target illumination prior under spherical harmonic lighting assumption.
Compared with the previous SOTA method in the portrait relighting task 
(\ie DPR~\cite{zhou2019deep}), our method designs a unique way for 
arbitrary illumination manipulation.
DPR~\cite{sun2019single} directly concatenates image features with 
lighting features from the encoder and feeds them into its decoder.
This way broadcasts the 1D lighting parameters 
to a 3D tensor whose size matches that of the image, 
which brings highly redundant and memory consumption. 
Considering the invariance of global illumination in a scene, the 
network should describe illumination in a 1D representation.
According to this intuition, 
we investigate the attention mechanism for lighting prior injection and 
progressively manipulate the illumination condition. 
To the best of our knowledge, our method is the first one which 
utilizes the attention mechanism to represent illumination conditions. 
Owing to the attention mechanism,
we maintain the lighting parameters in the 1D shape in the entire procedure, 
which achieves a better efficiency.

\subsection{Depth-guided geometry encoder}
\label{ssec:aux_enc}

Depth is important information for relighting task to make the network understand 3D dependencies.
In order to take fully advantages of depth, we further introduce surface normal derived from depth, 
which is strongly related to the local brightness of surface 
and the orientations of reflected light.
Besides, normal fetches detailed structural information and 
is conducive to local structure preservation.
We firstly select depth and normal as additional inputs 
which promote network to understand more complicated geometric 
and structural information for guiding relighting.
When only depth is given, we can calculate surface normal 
based on the following formulation:
\begin{equation}
    \begin{split}
    \vec{n}_{x, y} &= \frac{(\frac{\partial D_{x, y}}{\partial x}, \frac{\partial D_{x, y}}{\partial y}, -1)}{|\vec{n}_{x, y}|},\\
    &= \frac{(\frac{D_{x+1, y}-D_{x-1, y}}{2}, \frac{D_{x, y+1}-D_{x, y-1}}{2}, -1)}{|\vec{n}_{x, y}|}.
    \end{split}
\end{equation}

Besides, the convolution is shift-invariant, 
which means image patches at any position are treated equally, 
while the intensity of incident light depends on the global position 
in an image.
The convolution is blind to such global positional 
information~\cite{islam2020much, kayhan2020translation}.
To alleviate this problem, we utilize positional encoding,
which encodes global distance into local patches 
to assist the convolution to be aware of global distance.
Widely used sinusoidal positional encoding~\cite{vaswani2017attention} 
encodes positional information through sine and cosine functions 
with different periods.
It needs lots of channels and is memory-consuming 
in the high-resolution case. 
So we choose to use a light-weight linear positional encoding 
based on the Cartesian coordinate that encodes positional information 
by a 2D feature:
\begin{equation}
    PE(x,y)=2 \cdot [\frac{x}{W}, \frac{y}{H}] - 1.
\end{equation}
The value range of the linear positional encoding is $[-1,1]$.

\begin{figure}
    \centering
    \includegraphics[width=0.48\textwidth]{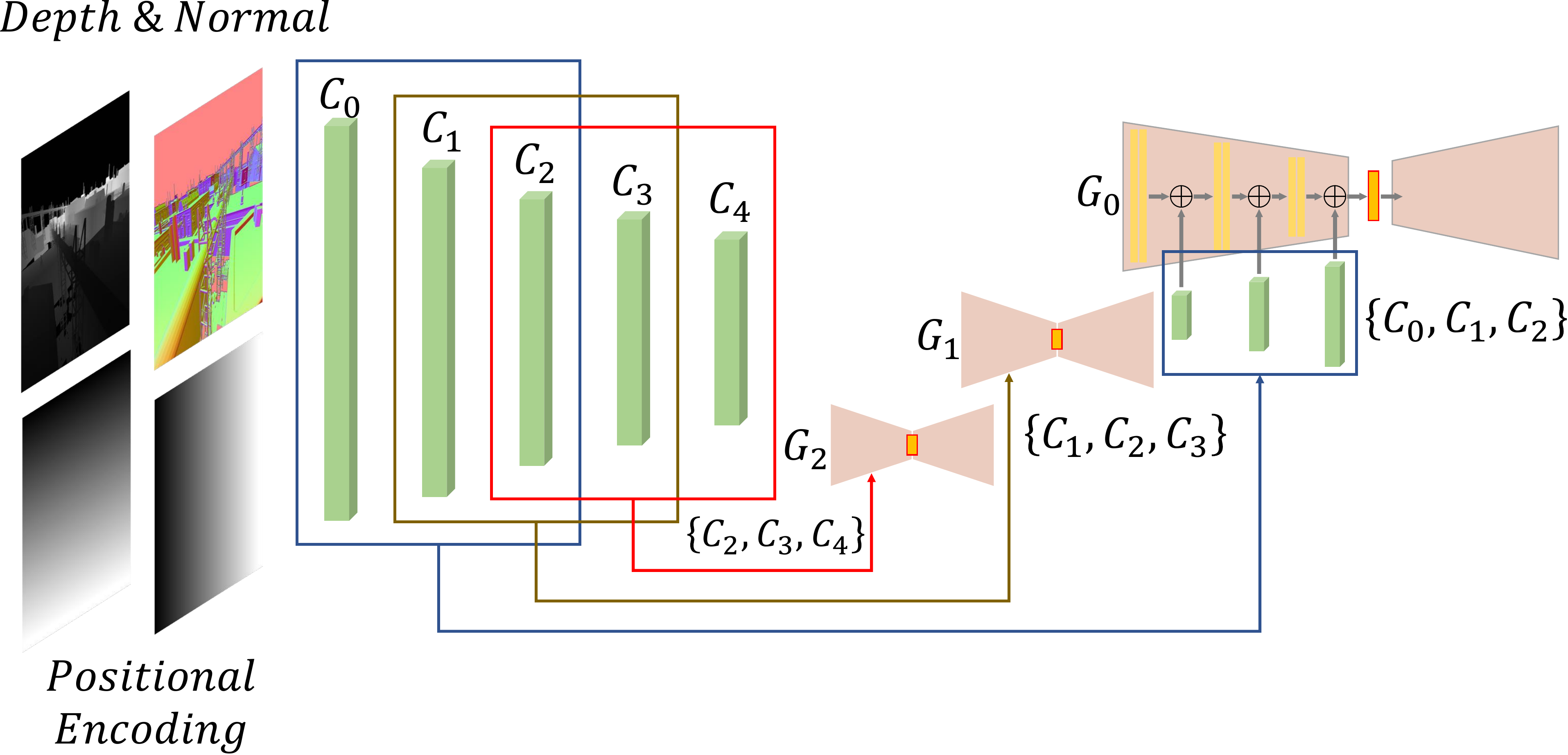}
    \caption{Illustration of depth-guided geometry encoder (DGGE). 
                DGGE has 5 levels in total and each level in the form of 
                ``[ReLU-Conv] $\times 2$''.
                This encoder is used to extract geometry- and 
                structure-related features from additional information.
                }
    \label{fig:aux_enc}
\end{figure}

We feed the above three guidance (\ie depth, normal, and positional encoding) 
into the proposed depth-guided geometry encoder (DGGE) 
to extract geometry- and structure-related features as illustrated in \figref{fig:aux_enc}, 
aiming to assist the network in understanding geometric relationships in scenes
and in recovering structure details in shadows.
To establish connections among levels for encoding the shared information, 
we design an encoder shared among levels. 
The DGGE has $5$ stages and provide $5$ intermediate feature maps in total.
We attempt to utilize as much information from depth as possible, 
so we densely merge features into the main stem. 
To achieve dense merging, $5$ intermediate feature maps are 
overlappingly divided into $3$ groups as 
$\{C^0, C^1, C^2\}, \{C^1, C^2, C^3\}, \{C^{2}, C^{3}, C^{4}\}$. 
Besides, this design is memory-efficient and suitable to relighting 
on the high resolution.
These groups of features correspond to $3$ levels of network.
Then they are merged with RGB features extracted by the encoders 
in the pyramid-like architecture:
\begin{equation}
    E^l_i = E^l_i + C^{l+i}.
\end{equation}
In this way, we enrich those RGB features by extracting 
geometry- and structure-related information.

\begin{figure}[h]
    \centering
    \includegraphics[width=0.48\textwidth]{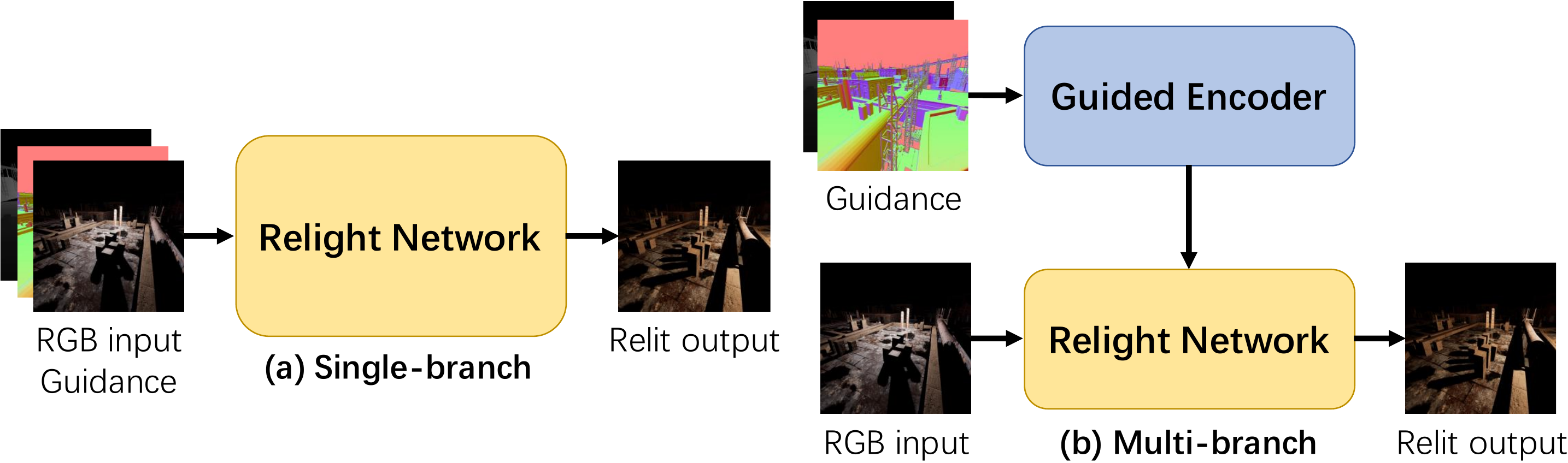}
    \caption{
     Two groups of methods in the respect of guidance usage. 
     The first group named `single-branch method'
      simply concatenates RGB images with guidance 
     and the second one named `multi-branch method' 
     individually encodes images and guidance. 
     Our method belongs to the second group.
    }
    \label{fig:2g}
\end{figure}

\noindent
\textbf{Discussions.} 
Basically, the existing relighting methods can be divided into two groups 
in the respect of guidance usage, as illustrated in~\figref{fig:2g}.
The first group of methods~\cite{helou2021ntire, wang2021multi} 
concatenate RGB images with guidance 
and feed them together into a single-branch network. 
However, they ignore the diversity 
in the distributions of RGB images and guided information, 
leading to inferior results. 
The second group of methods~\cite{wang2021multi} develop an additional branch 
to individually extract guided information.
Our method belongs to this group as well. Along with this idea, we make 
several modifications for a further improvement in performance. 
Except for depth, we detect two extra guided information 
(\ie surface normal and linear positional encoding). 
Surface normal derived from depth serves as an indispensable role in 
traditional rendering. 
Linear positional encoding considers 
the bias caused by diverse positions in the image 
while convolutions alone are unable to recognize such bias 
due to its shift-invariance.
Besides, we reveal that the finest information on the original resolution 
is crucial. 
Nevertheless, 
previous methods~\cite{wang2021multi} only fuse features after down-sampling. 
We also take efficiency into account. 
Previous methods (\eg MBNet~\cite{yang2021multi} and ADNet~\cite{helou2021ntire}) 
utilize complicated modules to conduct information fusion, 
which lead to highly redundant. 
Our method proves that simple addition operation is sufficient, 
which is friendly for high-resolution applications in practical usage.

\subsection{Implementation Details}

Our proposed network is jointly trained by optimizing reconstruction loss and grayscale 
SSIM loss to balance the fidelity of color and regional consistency of luminance.

\noindent
\textbf{Loss function}. 
SSIM loss~\cite{zhao2016loss} has been widely used in relighting task
~\cite{puthussery2020wdrn,hu2020sa,yang2021multi,yazdani2021physically} 
to enhance the structure consistency. 
Since L1 loss has already supervised color fidelity, we intend 
SSIM loss~\cite{zhao2016loss} to pay more attention 
to illumination consistency rather that color consistency. 
Thus, we further alter it to grayscale version as
\begin{equation}
    L_{SSIM}(\phi(I_{out}), \phi(I_{gt})) = 1 - SSIM(\phi(I_{out}), \phi(I_{gt})).
\end{equation}
where $\phi$ means the function that converts color image into grayscale one, $I_{out}$ is 
a relighting image and $I_{gt}$ is the ground truth.
Gradient loss are also utilized to yield sharper results~\cite{murmann2019dataset}, 
which can be formulated as
\begin{equation}
    \begin{aligned}
        L_{Gradient} &= \sum_x{\sum_y{\norm{\nabla I_{out}(x,y) - \nabla I_{gt}(x,y)}_2}}, \\
        \nabla I(x,y) &= (\frac{\partial I(x, y)}{\partial x}, \frac{\partial I(x, y)}{\partial y}), \\
        \frac{\partial I(x, y)}{\partial x} &= I(x+1, y) - I(x-1, y), \\
        \frac{\partial I(x, y)}{\partial y} &= I(x, y+1) - I(x, y-1),
    \end{aligned}
\end{equation}
where $I(x,y)$ means pixel value of $I$ at position $(x,y)$.

For the $l$-th level, the total loss is 
\begin{equation}
    \begin{aligned}
    L^l(I_{out}, I_{gt}) =& \alpha L_1(I_{out}, I_{gt}) + \beta L_{SSIM}(\phi(I_{out}), \phi(I_{gt})) \\
                &+ \gamma L_{Gradient}(I_{out}, I_{gt}).
    \end{aligned}
\end{equation}

All levels of the pyramid-like architecture utilize the same form of the loss function. 
So the final loss is 
\begin{equation}
    L = \sum_{l=0}^{2}{\mu^l L^l(I_{out}, I_{gt})}.
\end{equation}
We assign the same weight for each level, where $\mu^0=\mu^1=\mu^2=1.0$. 
The choice will be discussed in \secref{ssec:ab}.

\begin{table}[ht]
    \centering
    \renewcommand\arraystretch{1.2}
    \renewcommand{\tabcolsep}{3.4mm}
    \caption{Details of encoder, decoder, and light projection modules.}
    \begin{tabular}{c|c|c}
    \hline
        Module & Structure & Output size \\
    \hline
    \multirow{5}{*}{Encoder}    & $
                                \begin{bmatrix}
                                    (3\times 3, 48, 1) \& ReLU \\
                                \end{bmatrix}
                                \times 2$ &\multirow{1}{*}{$H\times W\times 48$}  \\
                    \cline{3-3} 
                        & $
                    \begin{bmatrix}
                        (3\times 3, 48, 2) \& ReLU \\
                    \end{bmatrix}
                    \times 1
                    $ &\multirow{2}{*}{$\frac{H}{2}\times \frac{W}{2}\times 48$}         \\
                        & $
                    \begin{bmatrix}
                        (3\times 3, 48, 1) \& ReLU \\
                    \end{bmatrix}
                    \times 1
                    $ &                                         \\
                    \cline{3-3} 
                        & $
                    \begin{bmatrix}
                        (3\times 3, 48, 2) \& ReLU \\
                    \end{bmatrix}
                    \times 1
                    $ & \multirow{5}{*}{$\frac{H}{4}\times \frac{W}{4}\times 48$}        \\
                        & $(3\times 3, 48, 1)$ &             \\
        \cline{1-2}
        Bottleneck   &  IARB $\times 4$ &                      \\
        \cline{1-2}
        \multirow{7}{*}{Decoder}     & $ReLU$ & \\
                        & $
                    \begin{bmatrix}
                        (3\times 3, 48, 2) \& ReLU \\
                    \end{bmatrix}
                    \times 3$ &                                     \\
                    \cline{3-3}  
                        & $Bilinear$ $2\times$ & \multirow{2}{*}{$\frac{H}{2}\times \frac{W}{2}\times 48$} \\
                        & $
                    \begin{bmatrix}
                        (3\times 3, 48, 1) \& ReLU \\
                    \end{bmatrix}
                    \times 3$ & \\
                    \cline{3-3}  
                        & $Bilinear$ $2\times$ & \multirow{2}{*}{$H\times W\times 48$} \\
                        & $
                    \begin{bmatrix}
                        (3\times 3, 48, 1) \& ReLU \\
                    \end{bmatrix}
                    \times 3$ &                 \\
                    \cline{3-3}
                        & $(3\times 3, 3, 1)$ & $H\times W\times 3$ \\
    \hline
    \hline
    \multirow{3}{*}{Projector} & $
    \begin{bmatrix}
        9\rightarrow 1024 \& ReLU \\
    \end{bmatrix}
    $ & \multirow{2}{*}{$1024$} \\
    & $
    \begin{bmatrix}
        1024\rightarrow 1024 \& ReLU \\
    \end{bmatrix}
    $ &                             \\
    \cline{3-3}
    &  $
    \begin{bmatrix}
        1024\rightarrow 12\times 192 \\
        
    \end{bmatrix}
    $ & $12\times 192$ \\
    \hline
    \end{tabular}
    \label{tab:arch}
\end{table}

\noindent
\textbf{Network details.}
In practice, our network has three levels and each level shares 
the same encoder-bottleneck-decoder structure, 
which has 2.67M parameters in total.
The detailed structure is presented in \tabref{tab:arch}, 
where $(k\times k, ch, s)$ represents a $k\times k$ convolution with stride $s$ 
whose output has $ch$ channels.
The bottleneck consists of 4 IARBs. In IARB, channels of each dilated convolution are 48 as well.
Note that IARB has convolutions with 3 different dilation rates, the intermediate channels are 144, 
and we utilize a $3\times 3$ convolution to reduce number of channels back to 48.
The structure of light projection module (\ie Projector) is also presented in \tabref{tab:arch}.
$X\rightarrow Y$ means a linear layer which project an $X$-dimension input to a $Y$-dimension output.

\section{Experiment}
\label{sec:exp}
\subsection{Datasets}
\label{ssec:data}
We train and test our proposed method on the Virtual Image Dataset for Illumination 
Transfer (\ie VIDIT~\cite{helou2020vidit}), which is utilized in AIM 2020~\cite{helou2021ntire} 
and NTIRE 2021~\cite{2020AIM} competitions.
It contains 15,600 images rendered by Unreal Engine 4 
which are captured from 390 virtual different outdoor scenes.
Miscellaneous objects with various surfaces and materials appear in VIDIT~\cite{helou2020vidit} 
dataset, such as metal, wood, stone, water, plant, fabric, smoke, fire, plastic, \etc. 
The illumination settings are all the combinations of 5 color temperatures (2500K, 3500K, 4500K,
5500K and 6500K) and 8 light directions (N, NE, E, SE, S, SW, W, NW).
The size of images is $1024\times 1024$ and corresponding depth maps of the same size 
are provided as well.
Similar with AIM 2020 and NTIRE 2021 competitions, we mainly force on 2 specific illumination settings 
($\theta_i=North,T_i=4500K; \theta_o=East, T_o=6500K$) for one-to-one relighting task.
Specifically, we convert images under illumination setting ($\theta_i=North,T_i=4500K$) 
to ones under another ($\theta_o=East, T_o=6500K$).
300 images in total are used for training and 45 images are used for validation. 
This setting is the same as when we participated the one-to-one relighting track~\cite{helou2021ntire} 
in NTIRE 2021 competition.
Under this setting, our team named 'MCG-NKU' achieves 
the best performance on the VIDIT~\cite{helou2020vidit} validation set.

Besides, we validate the proposed method on a dataset which captures from indoor scenes, 
\ie Multi-Illumination dataset~\cite{murmann2019dataset}.
This dataset consists of 1016 interior scenes 
in 95 different rooms throughout 12 residential and office buildings.
Each scene is filled with miscellaneous objects and clutter of various materials, 
which decorate in a typical domestic and office style.
All images are photographed under 25 pre-determined illumination settings.
25400 images whose size is $1500\times 1000$ and dense material labels segmented 
by crowd workers are provided.
HDR images obtained by merging exposures are furnished as well.
In our experiment, we take \textit{dir\_0} as input illumination settings 
and \textit{dir\_17} as output illumination settings in dataset 
to train all methods for evaluation.
On this dataset, 985 images are used for training and 30 images are used for validation.

We also conduct experiments on a domain-specific relighting dataset 
which is proposed by DPR~\cite{zhou2019deep} (\ie DPR dataset).
The DPR~\cite{zhou2019deep} dataset is built 
on the high-resolution CelebA~\cite{liu2015faceattributes} dataset (\ie CelebA-HQ) 
which contains 30,000 face images from the CelebA~\cite{liu2015faceattributes} dataset with 
size of $1024\times 1024$.
For each image, they randomly select 5 lighting conditions 
from a lighting prior dataset to generate relit face images, 
leading to 138,135 relit images. 
The authors of DPR~\cite{zhou2019deep} do not release their test dataset 
and detailed test setting, 
so we separate the images for the last 100 human faces 
under two diverse light conditions as test pairs. 
The remainder pairs are used for training in our experiment.

\subsection{Training details}

The parameters of our IAN are initialized by Xavier initialization~\cite{glorot2010understanding}.
We use Adam~\cite{Adam15} optimizer during training.

On VIDIT~\cite{helou2020vidit} dataset, we train the network for 24k iterations in total 
and utilize horizontal flip as data augmentation. 
Specifically, We select images of light source in the west and 
utilize horizontal flipping to fabricate images of light source in the east.
We directly feed full resolution images whose size is $1024\times1024$ into network.
The weights of losses are set to $\alpha=1.0, \beta=0.5, \gamma=0.0$.
For other comparison methods, 
we keep training settings in original papers for ones which conducted experiments 
on the VIDIT~\cite{helou2020vidit} dataset and use our training settings for ones which did not.
On Multi-Illumination~\cite{murmann2019dataset} dataset, we train our network for 120k iterations. 
We remove the DGGE when train on this dataset, for depth or normal information is unavailable. 
For a fair comparison, the weights of losses are set to $\alpha=1.0, \beta=0.0, \gamma=0.5$, 
following the setting for training the baseline 
on Multi-Illumination~\cite{murmann2019dataset} dataset.
Due to the limitation of GPU memory, 
we crop images to $992\times 992$ in training.
For fair comparison, all comparison methods on Multi-Illumination~\cite{murmann2019dataset} dataset 
are trained under the same setting.
Our IAN is trained under our setting as mentioned in \secref{ssec:data}.
Owing to the light projection module, 
one pretrained model of our IAN can manipulate arbitrary light condition.
We utilize both the official pretrained model and 
retrain DPR~\cite{zhou2019deep} under our setting for comparison.
Both methods are trained for 144k iterations on full resolution images, and 
we follow the loss setting in their paper~\cite{zhou2019deep} as $\alpha=1.0, \beta=0.0, \gamma=1.0$.

For all experiments, learning rate is set to $1e-4$ and batch size is set to $5$.
We train our IAN on one NVIDIA RTX TITAN and each 10k iterations
consume about 6 hours.

\subsection{Ablation studies}
\label{ssec:ab}
In this section, we show ablation studies we conduct to set forth the effectiveness 
of our method and give detailed analyses about the proposed modules.
We mainly investigate four factors influencing the performance of our proposed network.
We validate the effectiveness of the proposed illumination-aware residual block (IARB) firstly 
and then discuss the choice of loss weights.
Besides, various skip connections which are introduced in our work are experimented.
At last, diverse additional information fed to DGGE will be analyzed.
All ablation studies are conducted on the VIDIT~\cite{helou2020vidit} dataset. 

\noindent
\textbf{Number of levels}. 
\begin{figure*}[ht]
    \centering
    \includegraphics[width=0.95\textwidth]{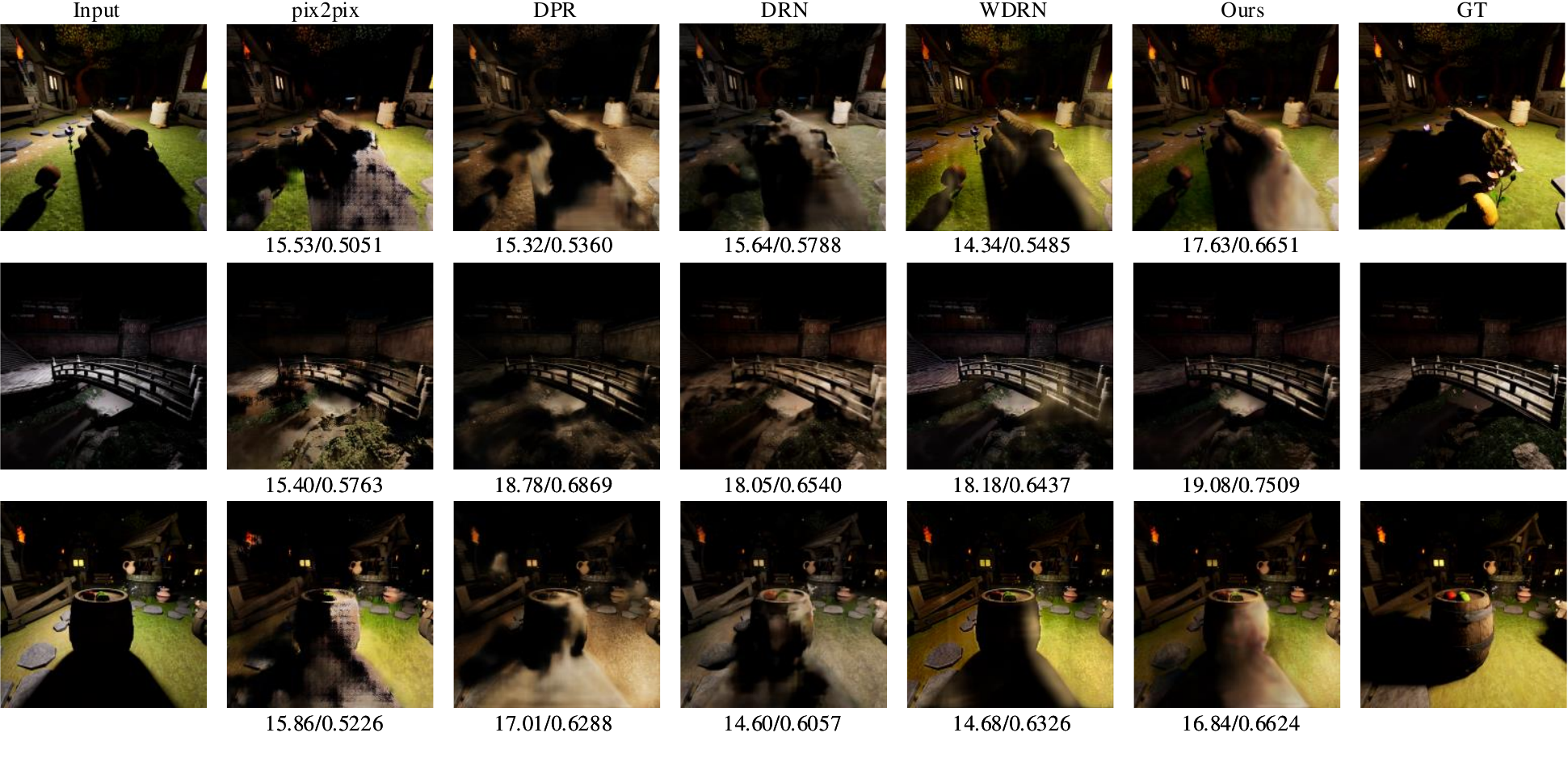}
    \caption{Representative results (w/o guidance). We select methods in AIM 2020~\cite{2020AIM}, 
            including DRN~\cite{wang2020deep}, WDRN~\cite{puthussery2020wdrn}. 
            We also select prestigious image-to-image translation method pix2pix~\cite{isola2017image} 
            and portrait relighting method DPR~\cite{zhou2019deep}.}
    \label{fig:rep_wog}
\end{figure*} 
\begin{figure*}[ht]
    \centering
    \includegraphics[width=0.95\textwidth]{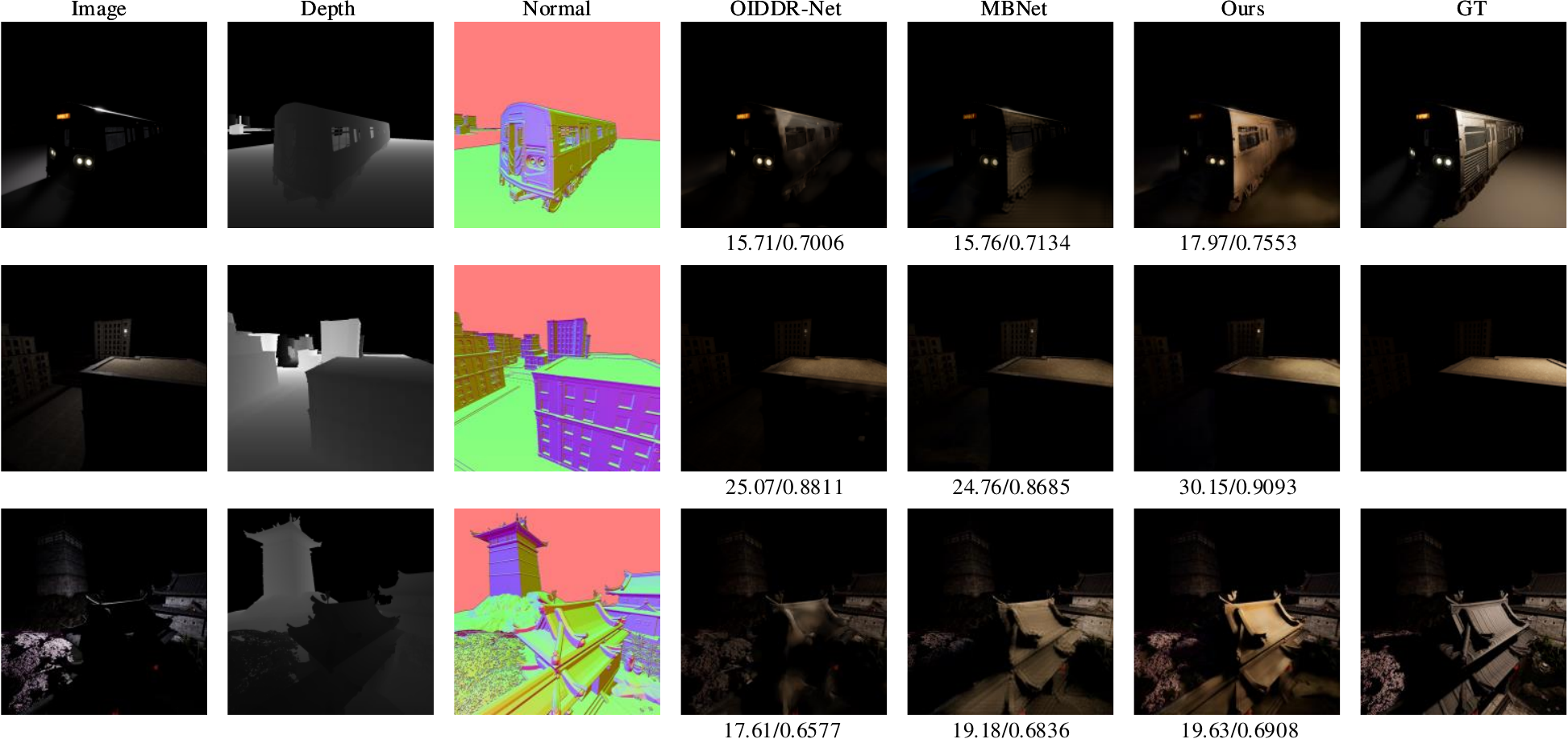}
    \caption{Representative results (w/ guidance). We compare our method with award-winning methods 
            in NTIRE 2021~\cite{helou2021ntire}, including MBNet~\cite{yang2021multi} 
            and OIDDR-Net~\cite{yazdani2021physically}.}
    \label{fig:rep_wg}
\end{figure*} 
Image pyramid is an effective tool to utilize multi-scale information. 
With the increase of levels, total calculating cost will slightly grow,
due to the decrease of image scale in each newer level.
However, introducing excessive levels brings redundant parameters and may 
lead to over-fitting problem. 
We thus examine how the number of pyramid level effects the performance of network 
and find a balance between efficiency and performance. 
We conduct experiments for $L=1,2,3,4$
and the results are shown in \tabref{tab:cmp_lvl}. 
\begin{table}[ht]
    \centering
    \renewcommand{\tabcolsep}{4.4mm}
    \caption{Quantitative evaluation for different levels.}
    \resizebox{0.48\textwidth}{9mm}{
    \begin{tabular}{l|c|c|c|c}
    \hline
        \makecell[c]{Levels} & PSNR & SSIM & Parameters & GMacs\\
    \hline
        $L=1$ & 19.19 & 0.7116 & 943.51k & 203.22 \\
        $L=2$ & 19.49 & 0.7213 & 1.81 M  & 245.43 \\
        $L=3$ & 19.70 & 0.7234 & 2.67 M  & 255.99 \\
        $L=4$ & 19.84 & 0.7300 & 3.53 M  & 258.63 \\
    \hline
    \end{tabular}}
    \label{tab:cmp_lvl}
\end{table}

As \tabref{tab:cmp_lvl} shown, performance continuously increases when the number of pyramid level grows.
However, quantities of parameters need to be stored when too many levels are involved in, 
which makes the network less efficient. 
So we finally set the number of pyramid levels $L$ to 3.

\noindent
\textbf{Choice of loss weights}.
The loss weights of three levels in the network are set individually.
In this part, we investigate how the choice of loss weights influences the final performance.

We first conduct experiments on loss weights among levels.
Intuitively, it is beneficial to increase loss weight with respect to the level, 
for the eventual aim of network is to obtain high-resolution relighting results 
on the finest level.
However, if we increase loss weights on finer levels too much, 
the network will degrade to a single large UNet-like network 
and loss its ability in progressive modeling. 
Contrarily, if we decrease loss weights, 
network undoubtedly outputs results ignored local details, 
which hinders the performance of the network.
We conduct experiments to prove above hypothesis. 
As shown in \tabref{tab:weight_lvls}, the results reveal that 
neither increasing nor decreasing is a good option. 
We finally select to assign the equal weight on each level, 
which shows the best performance in our experiments.
\begin{table}[ht]
    \centering
    \renewcommand{\tabcolsep}{4.6mm}
    \caption{Quantitative evaluations for loss weights among levels.}
    \begin{tabular}{c|c|c}
    \hline
        Zoom ratio & PSNR & SSIM \\
    \hline
        $10.0\times$  & 19.42 & 0.7031 \\
        $5.0\times$   & 19.52 & 0.7147 \\
        $1.0\times$   & 19.70 & 0.7234 \\
        $0.5\times$   & 19.53 & 0.7206 \\
        $0.1\times$   & 19.52 & 0.7206 \\
    \hline
    \end{tabular}
    \label{tab:weight_lvls}
\end{table}

\begin{figure*}[ht]
    \centering
    \includegraphics[width=0.95\textwidth]{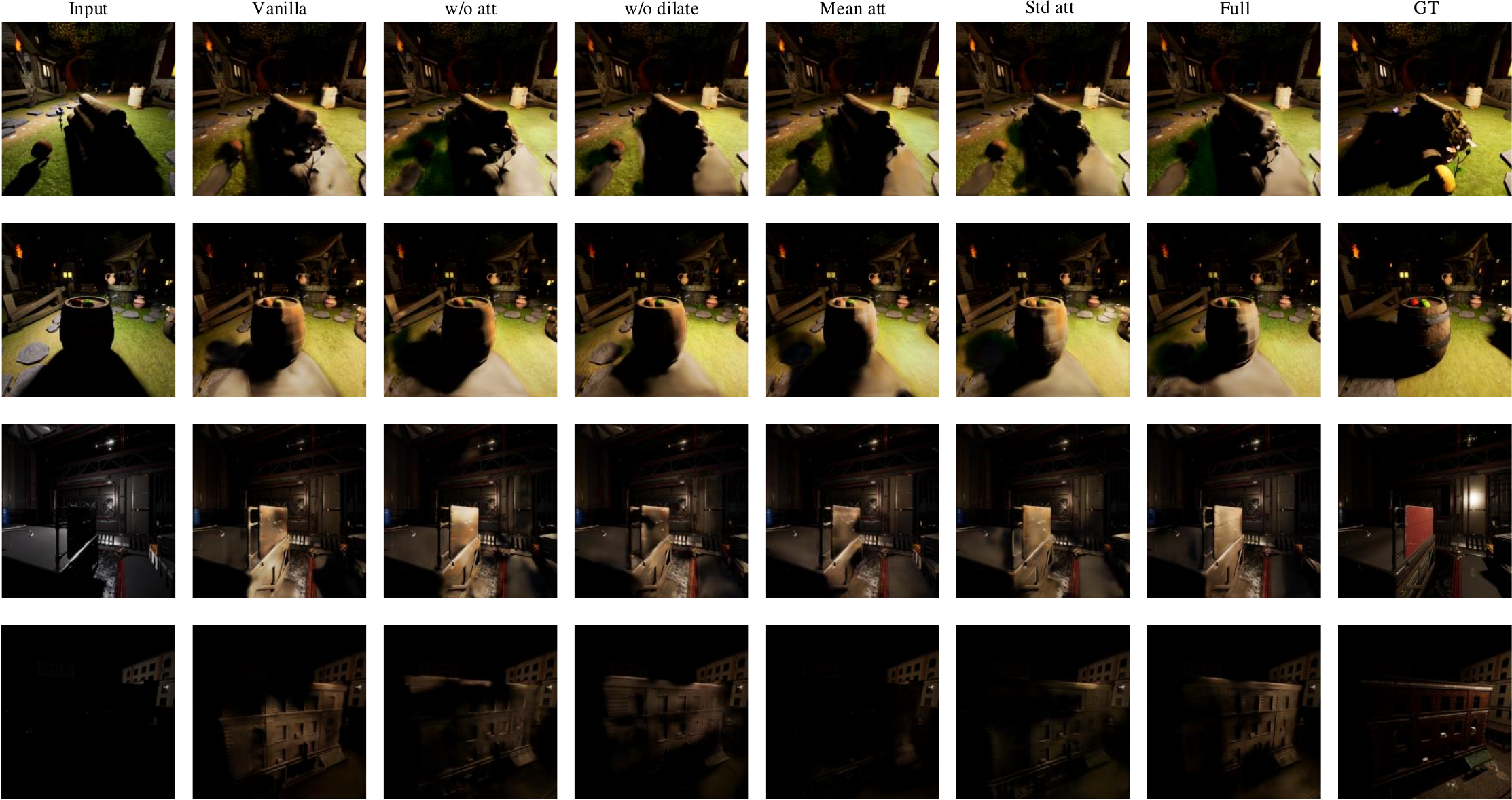}
    \caption{
    Qualitative comparisons between our IARB with other variants. 
    The results reveal the strong ability of proposed IARB in modeling light and relighting images.
    }
    \label{fig:ab_vis}
    \vspace{-5mm}
\end{figure*}

\noindent
\textbf{Investigation on IARB}. 
The proposed IARB serves as an important role in the IAN, 
which is designed based on the conventional ideas of rendering.
In this section, we confirm the effectiveness of components of IARB 
in \tabref{tab:cmp_res} and in \figref{fig:ab_vis}.
How the number of IARBs influences the performance is shown in \tabref{tab:cmp_num}.

\begin{table}[h]
    \centering
    \renewcommand{\tabcolsep}{4.6mm}
    \caption{Quantitative evaluations for IARB.}
    \begin{tabular}{c|c|c|c|c}
    \hline
    \multirow{2}{*}{Method} & \multicolumn{2}{c|}{w/ guidance} & \multicolumn{2}{c}{w/o guidance} \\
    \cline{2-5}
        & PSNR & SSIM & PSNR & SSIM \\
    \hline
        vanilla     & 19.24 & 0.7123 & 18.02 & 0.6717 \\
        w/o att     & 19.28 & 0.7184 & 18.05 & 0.6741 \\
        w/o dilated & 19.46 & 0.7208 & 18.16 & 0.6728 \\
        mean att    & 19.43 & 0.7223 & 18.17 & 0.6722 \\
        std att     & 19.62 & 0.7201 & 18.18 & 0.6527 \\
        full model  & 19.70 & 0.7234 & 18.27 & 0.6861 \\       
    \hline
    \end{tabular}
    \label{tab:cmp_res}
\end{table}

In \tabref{tab:cmp_res}, `vanilla' represents vanilla residual block~\cite{he2016deep}, 
`w/o att' means IARB without our statistical-coupled attention mechanism, 
`w/o dilated' denotes vanilla residual block with our proposed attention,
`mean att' represents dilated residual block with only mean attention 
and `std att' denotes dilated residual block with only standard deviation attention.

From \tabref{tab:cmp_res}, we can see that the performance decreases dramatically
without the proposed statistical-coupled attention mechanism,
indicating the significance of decoupling illumination descriptors. 
Only involving attention mechanism is also insufficient for 
acquiring accurate illumination descriptor under diverse frequencies.
Without distant samples obtained by dilated convolution, 
receptive field of the block is constrained and 
acquired descriptors thus become unreliable.
We also examine mean attention and standard deviation attention, respectively,
which proves the effectiveness of these illumination-related statistics.
In particular, standard deviation brings the most striking improvement, 
which indicates this statistic has a strong correlation 
with high frequent components of light as discussed in \secref{ssec:res}.

Besides, we display the qualitative results from these variants in \figref{fig:ab_vis}.
We can see that compared with other variants, our full model is superior in these aspects.
Among all results from the first and second row, 
only our full model can generate shadows in correct direction and position.
Without the usage of two-branch structure, 
the `vanilla' and `w/o att' module cannot extract precise light information. 
So we observe shadows from these variants are in wrong directions as shown in 
the first and second row.
Without dilated conventions, the `vanilla' and `w/o dilated' module fail to 
capture long-distance dependencies which are crucial for the consistency 
of illumination.
As a result, generated shadows or relit surfaces from these variants 
are hollow or broken.
Besides, only mean descriptor or deviation descriptor is insufficient to 
model the attributes of light source.
So surface in the results of `mean att' and `std att' 
which should be relit remains dark in the last row.
Owing to a large receptive field from dilated convolutions and 
precise modeling of light by diverse statistics, 
our method generates relighting images of high visual quality.
Through these examples, 
we reveal the strong ability of proposed IARB 
in modeling light and relighting images. 
Without our IARB, the ability of network in illumination-awareness is highly reduced, 
and it cannot extract precise attributes of light source for further light manipulation. 
\begin{table}[h]
    \centering
    \renewcommand{\tabcolsep}{4.6mm}
    \caption{Quantitative evaluation for different number of block(s).}
    \resizebox{0.48\textwidth}{12mm}{
    \begin{tabular}{l|c|c|c|c}
    \hline
        \makecell[c]{Levels} & PSNR & SSIM & Parameters & GMacs\\
    \hline
        $N=1$ & 19.09 & 0.7147 & 1.50 M & 223.76 \\
        $N=2$ & 19.28 & 0.7149 & 1.89 M & 234.50 \\
        $N=3$ & 19.51 & 0.7179 & 2.28 M & 245.24 \\
        $N=4$ & 19.70 & 0.7234 & 2.67 M & 255.99 \\
        $N=5$ & 19.74 & 0.7111 & 3.06 M & 266.73 \\
        $N=6$ & 19.71 & 0.7203 & 3.45 M & 277.47 \\
    \hline
    \end{tabular}}
    \label{tab:cmp_num}
\end{table}

Besides, we conduct experiments on number of IARBs in order to further 
prove the effectiveness of our proposed module and to find the optimal 
number in practical usage.
We experiment the number from 1 to 6 and the results is in \tabref{tab:cmp_num}.
While the number of IARBs increases, the performance increases correspondingly, 
which proves that IARB indeed benefits to relighting task.
However, due to inaccessibility of large mount of data, 
the performance reaches a plateau, 
which indicates network saturates after number of blocks exceeds 4. 
So we eventually select 4 blocks in our settings.

\noindent
\textbf{Investigation on skip connections}. 
\begin{figure}[ht]
    \centering
    \includegraphics[width=0.48\textwidth]{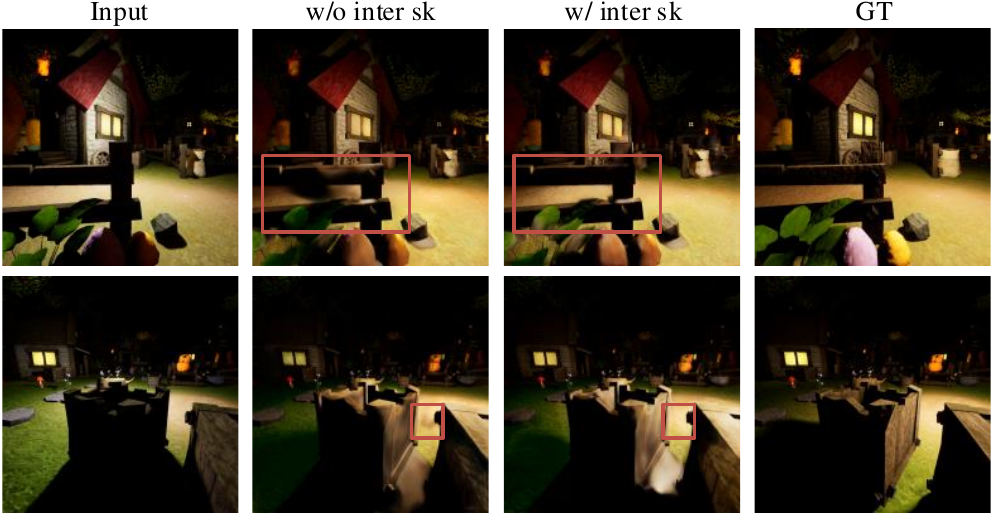}
    \caption{Once we remove cross-level skip connection, reconstruction information from high levels 
     is hard to transfer across levels, which makes network likely to generate images whose local and 
     global illumination is incoherent. 
    We highlight artifacts in images by red rectangle boxes.}
    \label{fig:inter}
\end{figure}
\begin{figure}
    \centering
    \includegraphics[width=0.48\textwidth]{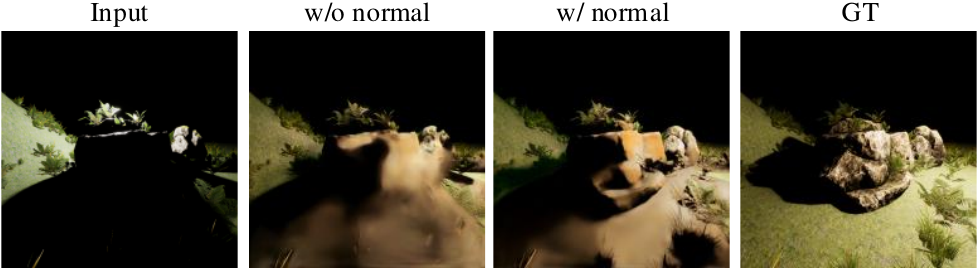}
    \caption{Normal is the most significant information among all additional information. Without it, 
            network cannot reconstruct detailed structures in dark environment.}
    \label{fig:norm}
\end{figure}
\begin{figure}[ht]
    \centering
    \includegraphics[width=0.48\textwidth]{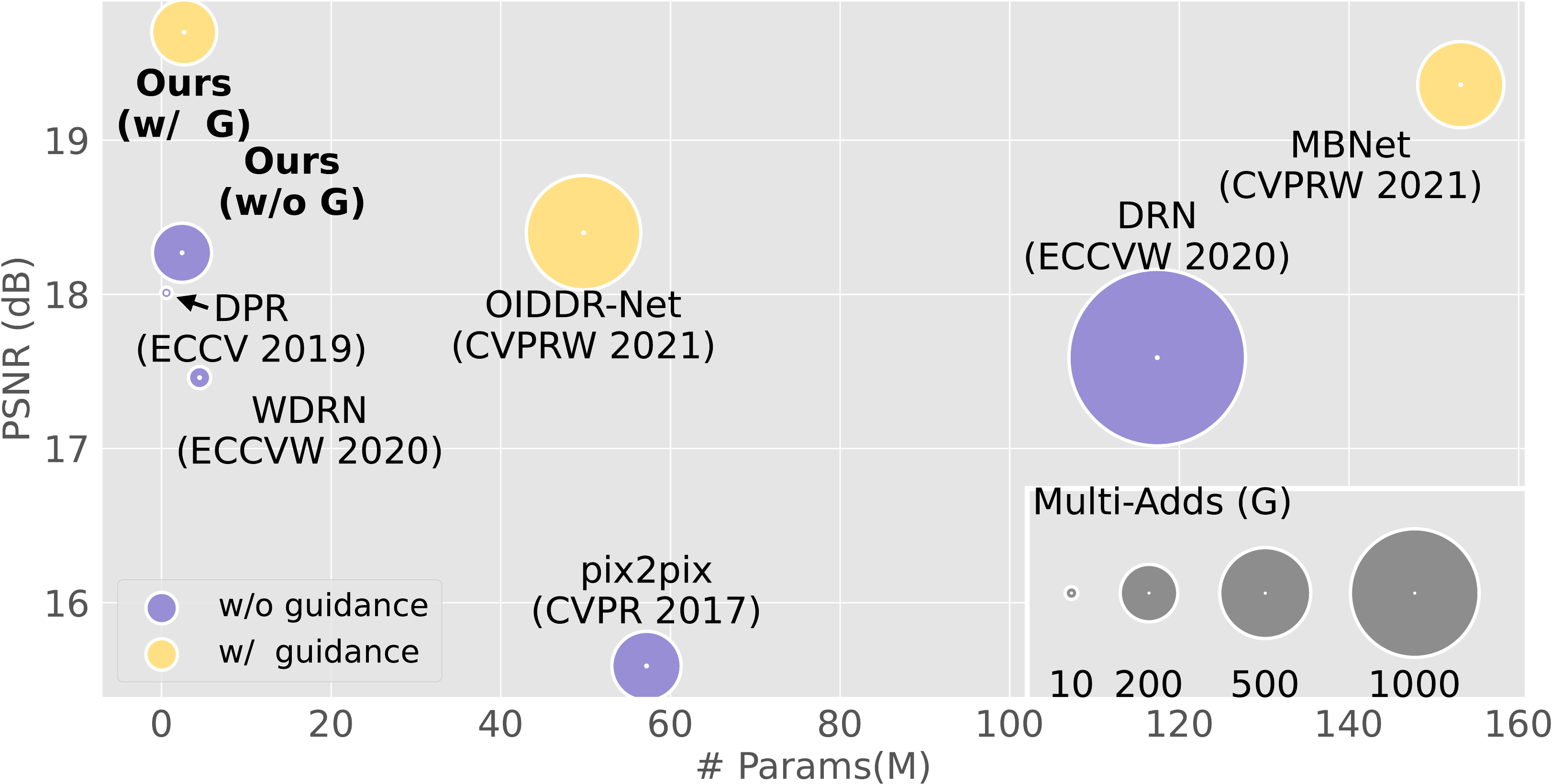}
    \caption{Comparison about performance, computational cost and number of parameters.}
    \label{fig:cplx}
    \vspace{-5mm}
\end{figure}
\begin{figure*}[ht]
    \centering
    \includegraphics[width=0.95\textwidth]{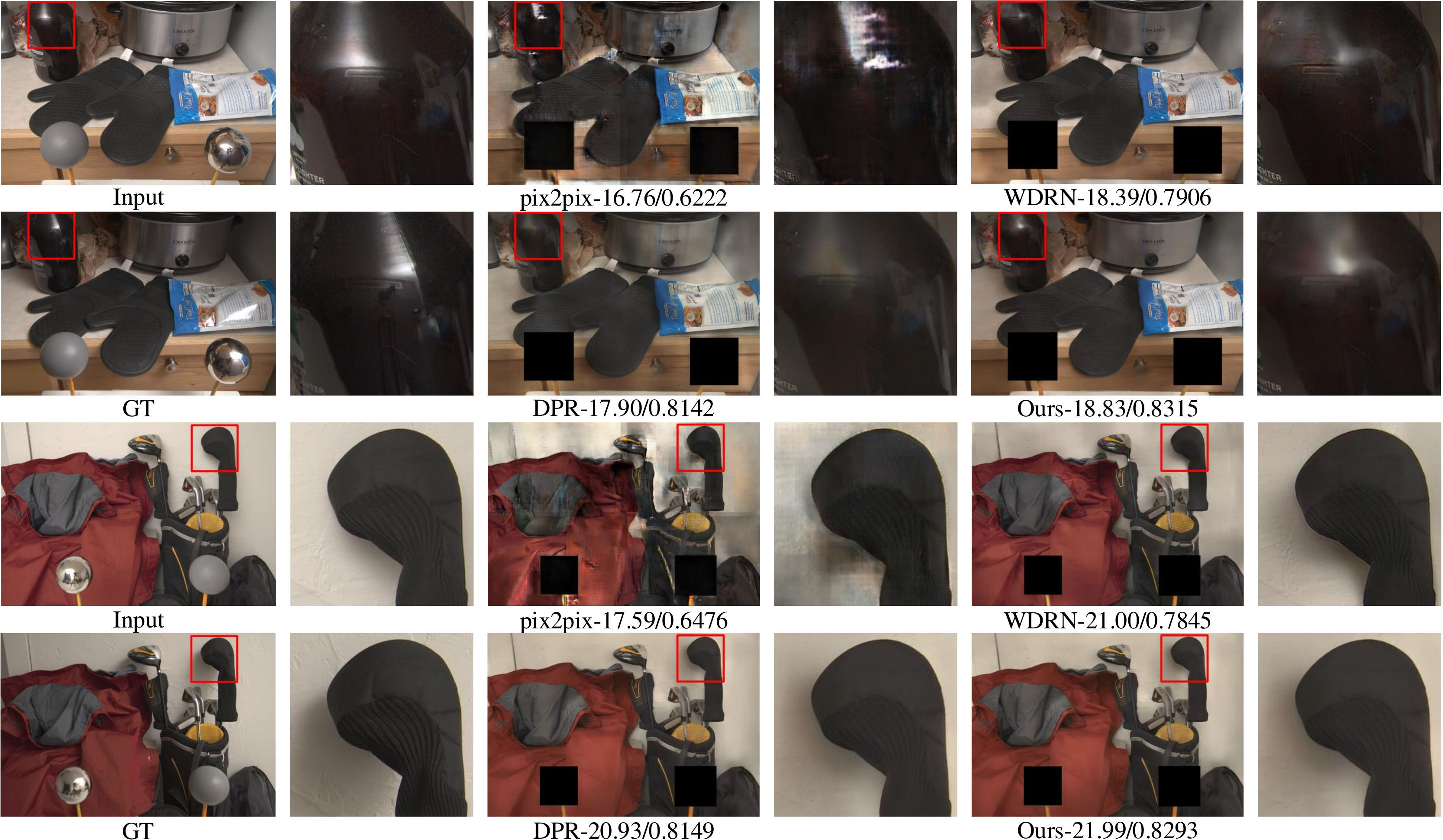}
    \caption{Representative results on multi-illumination dataset. 
            We present enlarged noteworthy patches in the lower right corner of overall images for detailed comparison.
            }
    \label{fig:rep_mi}
    \vspace{-5mm}
\end{figure*} 
\begin{figure*}[ht]
    \centering
    \includegraphics[width=0.95\textwidth]{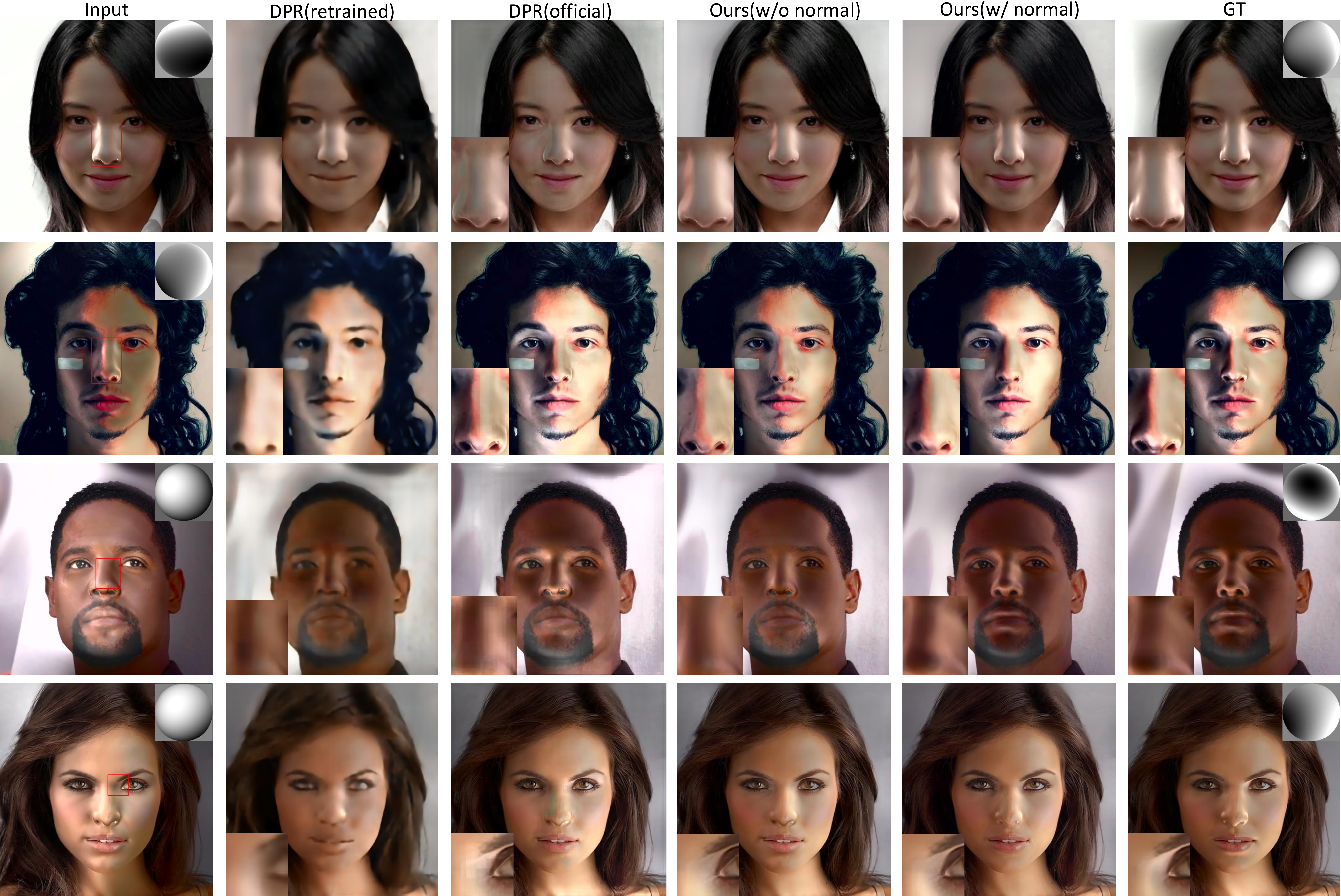}
    \caption{
    Qualitative comparison on the DPR~\cite{zhou2019deep} dataset. 
    Enlarged noteworthy patches are presented in the lower left corner of overall images for detailed comparison, 
    and visualizations of target light are presented in the upper right corner.
    }
    \label{fig:cmp_dpr}
    \vspace{-5mm}
\end{figure*} 
\begin{figure*}[ht]
    \centering
    \includegraphics[width=0.9\textwidth]{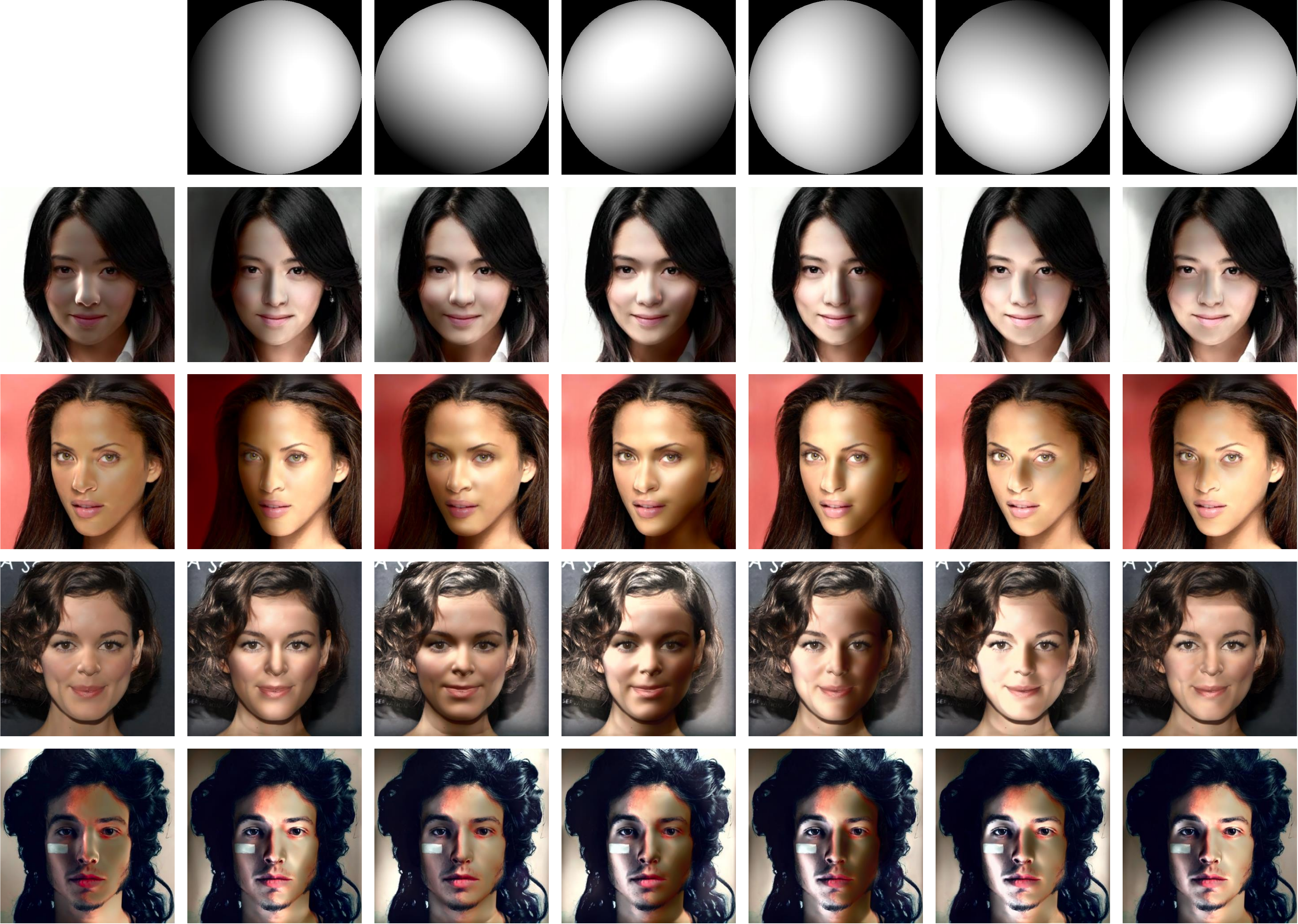}
    \caption{
        Qualitative results on the DPR~\cite{zhou2019deep} dataset with arbitrary light conditions. 
        The first row is visualizations of target spherical harmonic lights. 
        Input images in the first column are relit by above lights accordingly.
    }
    \label{fig:dyn_res}
    \vspace{-5mm}
\end{figure*} 
Previous works~\cite{puthussery2020wdrn, li2020multi, zhou2019deep} introduce 
skip connections aiming to ease the difficulty of training or to share 
information among scales.
We utilize various skip connections in our network as well.
In our work, three kinds of skip connections are introduced and  
named cross-level skip connection (CLSC), intra-level skip connection (ILSC), 
and image content skip connection (ICSC), respectively.

In this section, we conduct experiments about these skip connections 
and discuss their contributions in the proposed network.
\tabref{tab:sk} shows how objective metrics changes when we remove specific skip 
connections.
\begin{table}[h]
    \renewcommand{\tabcolsep}{4.6mm}
    \caption{Quantitative evaluation for diverse skip connections. 
    `w/ 'and `w/o' represent `with' and `without', respectively.}
    \centering
    \begin{tabular}{l|c|c}
    \hline
    \makecell[c]{Type} & PSNR & SSIM \\
    \hline
        full model   & 19.70 & 0.7234 \\   
        w/o ICSC     & 19.54 & 0.7104  \\
        w/o ILSC     & 19.49 & 0.7172  \\
        w/o CLSC     & 19.53 & 0.7159  \\
        w/o all SC   & 19.18 & 0.7027  \\
    \hline
    \end{tabular}
    \label{tab:sk}
\end{table}
As demonstrated in \secref{ssec:pyr}, CLSC enables the networks to be aware of global illumination 
when higher levels refine local details. 
Once we remove CLSC, we can observe conflicts between local details and global illumination, 
which leads to severe artifacts in relighting images as shown in \figref{fig:inter}.
ILSC and ICSC aim to transfer illumination invariant attributes to decoder side directly.
If we remove ILSC or ICSC connections, 
the detailed structures and object textures are hard to reconstruct, leading to the performance drop.

\noindent
\textbf{Investigation on depth-guided geometry encoder}. 
Depth-guided geometry encoder (DGGE) uses provided depth information, 
estimated surface normal and the positional encoding as input.
These additional information is evaluated as in \tabref{tab:cmp_aux} 
by removing depth, surface normal, and linear positional encoding, respectively.
Full model means utilizing all aforementioned guidance.
Bare model means the model that removes DGGE, indicating there is no additional input information.
\begin{table}[h]
    \centering
    \renewcommand{\tabcolsep}{4.6mm}
    \caption{Quantitative evaluation for different additional information in DGGE. 
    `lpe' means linear positional encoding.}
    \resizebox{0.48\textwidth}{10mm}{
    \begin{tabular}{l|c|c|c|c}
    \hline
    \makecell[c]{Structure} & PSNR & Diff & SSIM & Diff \\
    \hline
        full model  & 19.70 & ----- & 0.7234 & -----   \\
        w/o normal  & 18.33 & -1.37 & 0.6928 & -0.0306 \\
        w/o depth   & 19.57 & -0.13 & 0.7147 & -0.0087 \\
        w/o lpe     & 19.68 & -0.02 & 0.7006 & -0.0280 \\
        bare model  & 18.27 & -1.43 & 0.6861 & -0.0373 \\
    \hline
    \end{tabular}}
    \label{tab:cmp_aux}
\end{table}
Results reveal that surface normal brings the biggest performance gain 
among all additional information. 
Specifically, on the condition that we remove surface normal from DGGE, 
the performance of proposed network drops 1.37dB while performance 
only drops 0.13 dB if we remove depth map. 
In our observations, 
we find that removing normal results in blurring artifacts and ambiguous edges (see \figref{fig:norm}).
This phenomenon indicates that normal is an essential cue for local structure reconstruction. 
This result is consistent with the subjective metrics of results 
from the model guided by different information.

Besides, linear positional encoding is used as a spatial inductive bias to 
assist the network in capturing global structure with high fidelity. 
As shown in \tabref{tab:cmp_aux}, after removing positional encoding, 
PSNR drops slightly while SSIM~\cite{wang2004image} drops heavily, 
which indicates that positional encoding exerts the influence on structural consistency.

\subsection{Comparisons with state-of-the-art methods}
In this section, we compare our method with other state-of-the-art (SOTA) relighting methods.
The methods without depth guidance are mainly from AIM 2020~\cite{2020AIM} competition,
including WDRN~\cite{puthussery2020wdrn} which won the first position of AIM 
2020~\cite{2020AIM} and DRN~\cite{wang2020deep} which achieves the best PSNR score.
The methods with depth guidance are from NTIRE 2021~\cite{helou2021ntire} competition\footnote
{Results of NTIRE 2021 competition are available at 
\textit{https://competitions.codalab.org/competitions/28030\#results}
 and our team name is NK\_ZZL}, 
including MBNet~\cite{yang2021multi} which won the first position of NTIRE 2021~\cite{helou2021ntire} 
and OIDDR-Net~\cite{yazdani2021physically} which is the runner-up method.
Besides, we select pix2pix~\cite{isola2017image} which is a typical image-to-image translation 
method and DPR~\cite{zhou2019deep} which is a SOTA portrait relighting method for further comparison. 
For DPR~\cite{zhou2019deep}, we train a variant which removes light prediction module, 
because accurate light setting to train light prediction module is not provided in 
one-to-one relighting task.
For quantitative comparison, PSNR and SSIM~\cite{wang2004image} metrics 
are applied on RGB channel of relit results.
Moreover, the LPIPS metric~\cite{zhang2018unreasonable}, 
which is proven to be highly correlated with human ratings~\cite{jinjin2020pipal}, 
is also used for evaluation.

\noindent
\textbf{Efficiency.}
In this part, we give quantitative comparison about efficiency of relighting methods. 
Three main factors are selected for comparison, 
\ie performance, computational cost, and the number of parameters.
We use the number of composite 
multiply-accumulate operations~\cite{ahn2018fast} (Multi-Adds/Macs) for a single image 
as the measurement of computational cost.
We assume the input image size to be $1024\times 1024$ to calculate Multi-Adds.
This comparison is conducted on the VIDIT~\cite{helou2020vidit} dataset.
As illustrated in \figref{fig:cplx}, 
our proposed method uses relative few parameters and Macs to achieve the best performance 
than previous SOTA methods with or without additional guidance.

\noindent
\textbf{VIDIT Dataset.}
Compared with methods in AIM 2020~\cite{2020AIM} which lacks of additional guidance, 
our method outperforms them by a large margin in both 
distortion- and perception-oriented metrics.
With extra guidance, the performance of our proposed method still surpasses the
existing SOTA methods.
Except for comparisons on objective metrics, 
we also show representative results to compare perceptual quality of these methods 
and to further illustrate effectiveness of our method.
In \figref{fig:rep_wog}, 
we show the qualitative comparison results of methods without guidance.
Though network trained without geometric and structural guidance, 
we can see that 
it has the ability to reconstruct coarse object structure preliminarily 
as shown in $3$-rd row of \figref{fig:rep_wog}.
For these methods with guidance, the results are shown in \figref{fig:rep_wog}.
After utilizing guidance, 
our method can provide more consistent illumination and structural details 
with high fidelity as shown in the first and third row of \figref{fig:rep_wg}.
\begin{table}[h]
    \centering
    \renewcommand{\tabcolsep}{4.6mm}
    \caption{Quantitative evaluation on the VIDIT~\cite{helou2020vidit} dataset.}   
    \begin{tabular}{l|c|c|c}
    \hline
        \makecell[c]{Method} & PSNR$\uparrow$ & SSIM$\uparrow$ & LPIPS$\downarrow$\\
    \hline
        \multicolumn{4}{c}{w/o guidance} \\
    \hline
        pix2pix~\cite{isola2017image}          & 15.59 & 0.4890 & 0.4827  \\
        DRN~\cite{wang2020deep}                & 17.59 & 0.6151 & 0.3920  \\
        WDRN~\cite{puthussery2020wdrn}         & 17.46 & 0.6442 & 0.3299  \\
        DPR~\cite{zhou2019deep}                & 18.01 & 0.6389 & 0.3599  \\
        Ours (w/o guidance)                    & \textbf{18.27} & \textbf{0.6861} & \textbf{0.3077}  \\
    \hline
        \multicolumn{4}{c}{w/  guidance} \\
    \hline
        OIDDR-Net~\cite{yazdani2021physically} & 18.40 & 0.7039 & 0.2837  \\
        MBNet~\cite{yang2021multi}             & 19.36 & 0.7175 & 0.2928  \\
        Ours (w/ guidance)                     & \textbf{19.70} & \textbf{0.7234} & \textbf{0.2755}  \\
    \hline
    \end{tabular}
    \label{tab:cmp_sota}
\end{table}

\noindent
\textbf{Adobe Multi-Illumination Dataset.}
We further apply our experiments to a real scene dataset, \ie Adobe Multi-Illumination Dataset.
As mentioned in the paper~\cite{murmann2019dataset}, 
we mask chrome sphere and gray sphere 
which can be used as a prior to describe environment illumination 
during training and test phases.
Qualitative and quantitative results are shown in \tabref{tab:cmp_sota_mi} 
and \figref{fig:rep_mi}, respectively.
In \figref{fig:rep_mi},
the first and second row illustrate 
our method produces specular reflection on the surface of bottle with the best quality, 
which others either highlight wrong positions or cannot handle specular reflection at all.
The third and fourth row illustrate that our proposed method can generate more realistic shadows.
\tabref{tab:cmp_sota_mi} reveals quantitative comparison with previous methods and 
the performance of our proposed method is superior in the real scene dataset as well.

\begin{table}[h]
    \centering
    \renewcommand{\tabcolsep}{4.6mm}
    \caption{Quantitative evaluation on Adobe Multi-Illumination dataset. 
    Relighter is the official baseline which is proposed in ~\cite{murmann2019dataset}.}   
    \begin{tabular}{l|c|c|c}
    \hline
        \makecell[c]{Method} & PSNR$\uparrow$ & SSIM$\uparrow$ & LPIPS$\downarrow$\\
        \hline
        pix2pix~\cite{isola2017image}          & 17.46 & 0.6660 & 0.3597  \\
        DRN~\cite{wang2020deep}                & 14.40 & 0.6323 & 0.8004  \\
        WDRN~\cite{puthussery2020wdrn}         & 18.89 & 0.8160 & \textbf{0.1949}  \\
        DPR~\cite{zhou2019deep}                & 19.40 & 0.8377 & 0.2241  \\
        Relighter~\cite{murmann2019dataset}    & 16.54 & 0.7308 & 0.6261  \\
        Ours                                   & \textbf{19.67} & \textbf{0.8474} & 0.2100  \\
    \hline
    \end{tabular}
    \label{tab:cmp_sota_mi}
\end{table}

\noindent
\textbf{DPR Dataset.}
We extend our network to handle relighting task with arbitrary lighting condition 
and evaluate it on the DPR~\cite{zhou2019deep} dataset which is a portrait relighting dataset. 
To ensure the fairness, when compared with the previous SOTA method DPR~\cite{zhou2019deep}, 
we present both results from the official pretrained model (\ie DPR(official)) 
and them from the retrained model under our training setting (\ie DPR(retrained)).

\begin{table}[ht]
    \centering
    \renewcommand{\tabcolsep}{4.6mm}
    \caption{Quantitative evaluations on the DPR~\cite{zhou2019deep} dataset.}
    \begin{tabular}{l|c|c}
    \hline
    \makecell[c]{Type} & PSNR & SSIM \\
    \hline
        DPR (official)~\cite{zhou2019deep}   & 22.41 & 0.9189 \\
        DPR (retrained)~\cite{zhou2019deep}  & 24.00 & 0.8491 \\
        Ours (w/o normal)                    & 25.61 & 0.9509 \\
        Ours (w/  normal)                    & 27.98 & 0.9667 \\     
    \hline
    \end{tabular}
    \label{tab:cmp_dpr}
\end{table}
The results in \figref{fig:cmp_dpr} reveal that with simple modification, 
our method can tackle various light condition. 
When DPR~\cite{zhou2019deep} attempts to recast shadows, 
it frequently introduces obvious artifacts near edges of shadows, 
as the first to third rows shown. 
From the fourth row, we observe inconsistent effects of illumination 
in DPR~\cite{zhou2019deep}.
These results indicate that 
though DPR~\cite{zhou2019deep} supervises light attributes in spherical harmonic manner, 
it cannot fully take advantages of such explicit light condition. 
Instead, the results provided by our method 
have natural brightness change of high visual quality 
without obvious artifacts or undesirable sudden change of brightness. 
These results prove that 
implicitly injecting light condition into network is a better idea, 
which reduces explicit accumulation of errors.
Besides, we provide visual results under various light condition in \figref{fig:dyn_res}. 
By only modified the illumination branch, the network enables relighting for 
arbitrary light condition, 
which proves that our IARB can faithfully extract illumination-related information 
and this descriptor is effective when re-render the input image. 
The quantitative results shown in \tabref{tab:cmp_dpr}.
Both PSNR and SSIM values of our method surpass DPR~\cite{zhou2019deep} by a large margin, 
which reveals the effectiveness of our network.
In summary, 
our design purpose is consistent with the final results, and it is proven that 
the IARB is suitable for relighting task.

\section{Conclusion and Prospect}
In this paper, 
we thoroughly investigate previous relighting methods from diverse viewpoints 
and get inspirations from ideas of the conventional physical based rendering.
According to these inspirations, we design an illumination-aware network intrinsically 
suitable to the relighting task and deploy an illumination-aware residual block 
which approximates conventional rendering process to assist relighting.
Besides, we employ a depth-guided geometry encoder 
and utilize additional information beyond RGB images 
to acquire geometry- and structure-related information 
which benefits to relighting.
Adequate comparisons with previous SOTA methods and ablation studies 
reveal the effectiveness and efficiency of our proposed method.

However, there exists room for future improvement. 
Based on our observations in practice, we list the following aspects that should be emphasized in future work.
\begin{itemize}
    \item Relighting specular objects and transparent objects to be more realistic (see \figref{fig:rep_mi} Row 1-2).
    \item Completing textures of relighting regions according to surrounding patches or global style (see \figref{fig:rep_wg} Row 1-3).
    \item Utilizing inaccurate or sparse guidance which is more practical in reality to generate comparable results. 
\end{itemize}

\bibliographystyle{IEEEtran}
\bibliography{IAN}

\begin{thebibliography}{10}
\providecommand{\url}[1]{#1}
\csname url@samestyle\endcsname
\providecommand{\newblock}{\relax}
\providecommand{\bibinfo}[2]{#2}
\providecommand{\BIBentrySTDinterwordspacing}{\spaceskip=0pt\relax}
\providecommand{\BIBentryALTinterwordstretchfactor}{4}
\providecommand{\BIBentryALTinterwordspacing}{\spaceskip=\fontdimen2\font plus
\BIBentryALTinterwordstretchfactor\fontdimen3\font minus
  \fontdimen4\font\relax}
\providecommand{\BIBforeignlanguage}[2]{{%
\expandafter\ifx\csname l@#1\endcsname\relax
\typeout{** WARNING: IEEEtran.bst: No hyphenation pattern has been}%
\typeout{** loaded for the language `#1'. Using the pattern for}%
\typeout{** the default language instead.}%
\else
\language=\csname l@#1\endcsname
\fi
#2}}
\providecommand{\BIBdecl}{\relax}
\BIBdecl

\bibitem{sen2005dual}
P.~Sen, B.~Chen, G.~Garg, S.~R. Marschner, M.~Horowitz, M.~Levoy, and H.~P.
  Lensch, ``Dual photography,'' \emph{ACM Trans. Graph.}, 2005.

\bibitem{li2018learning}
Z.~Li, Z.~Xu, R.~Ramamoorthi, K.~Sunkavalli, and M.~Chandraker, ``Learning to
  reconstruct shape and spatially-varying reflectance from a single image,''
  \emph{ACM Trans. Graph.}, 2018.

\bibitem{xu2018deep}
Z.~Xu, K.~Sunkavalli, S.~Hadap, and R.~Ramamoorthi, ``Deep image-based
  relighting from optimal sparse samples,'' \emph{ACM Trans. Graph.}, 2018.

\bibitem{zhou2019deep}
H.~Zhou, S.~Hadap, K.~Sunkavalli, and D.~W. Jacobs, ``Deep single-image
  portrait relighting,'' in \emph{Int. Conf. Comput. Vis.}, 2019.

\bibitem{sun2019single}
T.~Sun, J.~T. Barron, Y.-T. Tsai, Z.~Xu, X.~Yu, G.~Fyffe, C.~Rhemann, J.~Busch,
  P.~E. Debevec, and R.~Ramamoorthi, ``Single image portrait relighting,''
  \emph{ACM Trans. Graph.}, 2019.

\bibitem{nestmeyer2020learning}
T.~Nestmeyer, J.-F. Lalonde, I.~Matthews, and A.~Lehrmann, ``Learning
  physics-guided face relighting under directional light,'' in \emph{IEEE Conf.
  Comput. Vis. Pattern Recog.}, 2020.

\bibitem{2020AIM}
M.~E. Helou, R.~Zhou, S.~Süsstrunk, R.~Timofte, and J.~Cheng, ``Aim 2020:
  Scene relighting and illumination estimation challenge,'' in \emph{Eur. Conf.
  Comput. Vis. Worksh.}, 2020.

\bibitem{7792614}
Z.~Han, J.~Tian, L.~Qu, and Y.~Tang, ``A new intrinsic-lighting color space for
  daytime outdoor images,'' \emph{IEEE Trans. Image Process.}, vol.~26, no.~2,
  pp. 1031--1039, 2017.

\bibitem{9725240}
F.~Zhan, Y.~Yu, C.~Zhang, R.~Wu, W.~Hu, S.~Lu, F.~Ma, X.~Xie, and L.~Shao,
  ``Gmlight: Lighting estimation via geometric distribution approximation,''
  \emph{IEEE Trans. Image Process.}, vol.~31, pp. 2268--2278, 2022.

\bibitem{9785513}
Z.~Hu, N.~E. Nsampi, X.~Wang, and Q.~Wang, ``Pnrnet: Physically-inspired neural
  rendering for any-to-any relighting,'' \emph{IEEE Trans. Image Process.},
  vol.~31, pp. 3935--3948, 2022.

\bibitem{7983410}
S.~Liu and M.~N. Do, ``Inverse rendering and relighting from multiple color
  plus depth images,'' \emph{IEEE Trans. Image Process.}, vol.~26, no.~10, pp.
  4951--4961, 2017.

\bibitem{lecun2015deep}
Y.~LeCun, Y.~Bengio, and G.~Hinton, ``Deep learning,'' \emph{Nature}, 2015.

\bibitem{simonyan2014very}
K.~Simonyan and A.~Zisserman, ``Very deep convolutional networks for
  large-scale image recognition,'' in \emph{Int. Conf. Learn. Represent.},
  2015.

\bibitem{he2016deep}
K.~He, X.~Zhang, S.~Ren, and J.~Sun, ``Deep residual learning for image
  recognition,'' in \emph{IEEE Conf. Comput. Vis. Pattern Recog.}, 2016.

\bibitem{sengupta2019neural}
S.~Sengupta, J.~Gu, K.~Kim, G.~Liu, D.~W. Jacobs, and J.~Kautz, ``Neural
  inverse rendering of an indoor scene from a single image,'' in \emph{Int.
  Conf. Comput. Vis.}, 2019.

\bibitem{Philip2019Multi}
J.~Philip, M.~Gharbi, T.~Zhou, A.~A. Efros, and G.~Drettakis, ``Multi-view
  relighting using a geometry-aware network,'' \emph{ACM Trans. Graph.}, 2019.

\bibitem{wang2020single}
Z.~Wang, X.~Yu, M.~Lu, Q.~Wang, C.~Qian, and F.~Xu, ``Single image portrait
  relighting via explicit multiple reflectance channel modeling,'' \emph{ACM
  Trans. Graph.}, 2020.

\bibitem{qiu2020towards}
D.~Qiu, J.~Zeng, Z.~Ke, W.~Sun, and C.~Yang, ``Towards geometry guided neural
  relighting with flash photography,'' \emph{arXiv preprint arXiv:2008.05157},
  2020.

\bibitem{yazdani2021physically}
A.~Yazdani, T.~Guo, and V.~Monga, ``Physically inspired dense fusion networks
  for relighting,'' in \emph{IEEE Conf. Comput. Vis. Pattern Recog. Worksh.},
  2021.

\bibitem{srinivasan2020nerv}
P.~P. Srinivasan, B.~Deng, X.~Zhang, M.~Tancik, B.~Mildenhall, and J.~T.
  Barron, ``Nerv: Neural reflectance and visibility fields for relighting and
  view synthesis,'' \emph{Eur. Conf. Comput. Vis.}, 2020.

\bibitem{niemeyer2021giraffe}
M.~Niemeyer and A.~Geiger, ``Giraffe: Representing scenes as compositional
  generative neural feature fields,'' in \emph{IEEE Conf. Comput. Vis. Pattern
  Recog.}, 2021.

\bibitem{schwarz2020graf}
K.~Schwarz, Y.~Liao, M.~Niemeyer, and A.~Geiger, ``Graf: Generative radiance
  fields for 3d-aware image synthesis,'' \emph{arXiv preprint
  arXiv:2007.02442}, 2020.

\bibitem{murmann2019dataset}
L.~Murmann, M.~Gharbi, M.~Aittala, and F.~Durand, ``A dataset of
  multi-illumination images in the wild,'' in \emph{Int. Conf. Comput. Vis.},
  2019.

\bibitem{helou2020vidit}
M.~E. Helou, R.~Zhou, J.~Barthas, and S.~S{\"u}sstrunk, ``Vidit: Virtual image
  dataset for illumination transfer,'' \emph{arXiv preprint arXiv:2005.05460},
  2020.

\bibitem{helou2021ntire}
M.~E. Helou, R.~Zhou, S.~Susstrunk, and R.~Timofte, ``Ntire 2021 depth guided
  image relighting challenge,'' \emph{IEEE Conf. Comput. Vis. Pattern Recog.
  Worksh.}, 2021.

\bibitem{ramamoorthi2001signal}
R.~Ramamoorthi and P.~Hanrahan, ``A signal-processing framework for inverse
  rendering,'' in \emph{SIGGRAPH}, 2001.

\bibitem{aldrian2012inverse}
O.~Aldrian and W.~A. Smith, ``Inverse rendering of faces with a 3d morphable
  model,'' \emph{IEEE Trans. Pattern Anal. Mach. Intell.}, 2012.

\bibitem{levoy1990efficient}
M.~Levoy, ``Efficient ray tracing of volume data,'' \emph{ACM Trans. Graph.},
  1990.

\bibitem{green2003spherical}
R.~Green, ``Spherical harmonic lighting: The gritty details,'' in
  \emph{Archives of the game developers conference}, 2003.

\bibitem{imageworks2010physically}
S.~P. Imageworks, ``Physically-based shading models in film and game
  production,'' 2010.

\bibitem{basri2003lambertian}
R.~Basri and D.~W. Jacobs, ``Lambertian reflectance and linear subspaces,''
  \emph{IEEE Trans. Pattern Anal. Mach. Intell.}

\bibitem{belhumeur1998set}
P.~N. Belhumeur and D.~J. Kriegman, ``What is the set of images of an object
  under all possible illumination conditions?'' \emph{Int. J. Comput. Vis.},
  1998.

\bibitem{ramamoorthi2001relationship}
R.~Ramamoorthi and P.~Hanrahan, ``On the relationship between radiance and
  irradiance: determining the illumination from images of a convex lambertian
  object,'' \emph{J. Opt. Soc. Am. A}, 2001.

\bibitem{debevec2000acquiring}
P.~Debevec, T.~Hawkins, C.~Tchou, H.-P. Duiker, W.~Sarokin, and M.~Sagar,
  ``Acquiring the reflectance field of a human face,'' in \emph{SIGGRAPH},
  2000.

\bibitem{wang2009kernel}
J.~Wang, Y.~Dong, X.~Tong, Z.~Lin, and B.~Guo, ``Kernel nystr{\"o}m method for
  light transport,'' in \emph{ACM Trans. Graph.}, 2009.

\bibitem{reddy2012frequency}
D.~Reddy, R.~Ramamoorthi, and B.~Curless, ``Frequency-space decomposition and
  acquisition of light transport under spatially varying illumination,'' in
  \emph{Eur. Conf. Comput. Vis.}, 2012.

\bibitem{malzbender2001polynomial}
T.~Malzbender, D.~Gelb, and H.~Wolters, ``Polynomial texture maps,'' in
  \emph{SIGGRAPH}, 2001.

\bibitem{karsch2011rendering}
K.~Karsch, V.~Hedau, D.~Forsyth, and D.~Hoiem, ``Rendering synthetic objects
  into legacy photographs,'' \emph{ACM Trans. Graph.}, 2011.

\bibitem{duchene2015multi}
S.~Duch{\^e}ne, C.~Riant, G.~Chaurasia, J.~Lopez-Moreno, P.-Y. Laffont,
  S.~Popov, A.~Bousseau, and G.~Drettakis, ``Multi-view intrinsic images of
  outdoors scenes with an application to relighting,'' \emph{ACM Trans.
  Graph.}, 2015.

\bibitem{zhang2016emptying}
E.~Zhang, M.~F. Cohen, and B.~Curless, ``Emptying, refurnishing, and relighting
  indoor spaces,'' \emph{ACM Trans. Graph.}, 2016.

\bibitem{krizhevsky2012imagenet}
A.~Krizhevsky, I.~Sutskever, and G.~E. Hinton, ``Imagenet classification with
  deep convolutional neural networks,'' \emph{Adv. Neural Inform. Process.
  Syst.}, 2012.

\bibitem{ding2019argan}
B.~Ding, C.~Long, L.~Zhang, and C.~Xiao, ``Argan: Attentive recurrent
  generative adversarial network for shadow detection and removal,'' in
  \emph{Int. Conf. Comput. Vis.}, 2019.

\bibitem{nagano2019deep}
K.~Nagano, H.~Luo, Z.~Wang, J.~Seo, J.~Xing, L.~Hu, L.~Wei, and H.~Li, ``Deep
  face normalization,'' \emph{ACM Trans. Graph.}, 2019.

\bibitem{zhang2020portrait}
X.~Zhang, J.~T. Barron, Y.-T. Tsai, R.~Pandey, X.~Zhang, R.~Ng, and D.~E.
  Jacobs, ``Portrait shadow manipulation,'' \emph{ACM Trans. Graph.}, 2020.

\bibitem{zhang2020copy}
Y.~Zhang, I.~W. Tsang, Y.~Luo, C.-H. Hu, X.~Lu, and X.~Yu, ``Copy and paste
  gan: Face hallucination from shaded thumbnails,'' in \emph{IEEE Conf. Comput.
  Vis. Pattern Recog.}, 2020.

\bibitem{ren2015image}
P.~Ren, Y.~Dong, S.~Lin, X.~Tong, and B.~Guo, ``Image based relighting using
  neural networks,'' \emph{ACM Trans. Graph.}, 2015.

\bibitem{yu2020self}
Y.~Yu, A.~Meka, M.~Elgharib, H.-P. Seidel, C.~Theobalt, and W.~A. Smith,
  ``Self-supervised outdoor scene relighting,'' in \emph{Eur. Conf. Comput.
  Vis.}, 2020.

\bibitem{mildenhall2020nerf}
B.~Mildenhall, P.~P. Srinivasan, M.~Tancik, J.~T. Barron, R.~Ramamoorthi, and
  R.~Ng, ``Nerf: Representing scenes as neural radiance fields for view
  synthesis,'' in \emph{Eur. Conf. Comput. Vis.}, 2020.

\bibitem{martin2021nerf}
R.~Martin-Brualla, N.~Radwan, M.~S. Sajjadi, J.~T. Barron, A.~Dosovitskiy, and
  D.~Duckworth, ``Nerf in the wild: Neural radiance fields for unconstrained
  photo collections,'' in \emph{IEEE Conf. Comput. Vis. Pattern Recog.}, 2021.

\bibitem{puthussery2020wdrn}
D.~Puthussery, M.~Kuriakose, J.~C~V \emph{et~al.}, ``Wdrn: A wavelet decomposed
  relightnet for image relighting,'' \emph{IEEE Conf. Comput. Vis. Pattern
  Recog. Worksh.}, 2020.

\bibitem{gafton20202d}
P.~Gafton and E.~Maraz, ``2d image relighting with image-to-image
  translation,'' \emph{arXiv preprint arXiv:2006.07816}, 2020.

\bibitem{isola2017image}
P.~Isola, J.-Y. Zhu, T.~Zhou, and A.~A. Efros, ``Image-to-image translation
  with conditional adversarial networks,'' in \emph{IEEE Conf. Comput. Vis.
  Pattern Recog.}, 2017.

\bibitem{yang2021multi}
H.-H. Yang, W.-T. Chen, H.-L. Luo, and S.-Y. Kuo, ``Multi-modal bifurcated
  network for depth guided image relighting,'' in \emph{IEEE Conf. Comput. Vis.
  Pattern Recog. Worksh.}, 2021.

\bibitem{pang2020hierarchical}
Y.~Pang, L.~Zhang, X.~Zhao, and H.~Lu, ``Hierarchical dynamic filtering network
  for rgb-d salient object detection,'' in \emph{Eur. Conf. Comput. Vis.},
  2020.

\bibitem{zhang2019deep}
H.~Zhang, Y.~Dai, H.~Li, and P.~Koniusz, ``Deep stacked hierarchical
  multi-patch network for image deblurring,'' in \emph{IEEE Conf. Comput. Vis.
  Pattern Recog.}, 2019.

\bibitem{das2020fast}
S.~D. Das and S.~Dutta, ``Fast deep multi-patch hierarchical network for
  nonhomogeneous image dehazing,'' in \emph{IEEE Conf. Comput. Vis. Pattern
  Recog. Worksh.}, 2020.

\bibitem{ronneberger2015u}
O.~Ronneberger, P.~Fischer, and T.~Brox, ``U-net: Convolutional networks for
  biomedical image segmentation,'' in \emph{Med. Image. Comput. Comput. Assist.
  Interv.}, 2015.

\bibitem{chen2020neural}
Z.~Chen, A.~Chen, G.~Zhang, C.~Wang, Y.~Ji, K.~N. Kutulakos, and J.~Yu, ``A
  neural rendering framework for free-viewpoint relighting,'' in \emph{IEEE
  Conf. Comput. Vis. Pattern Recog.}, 2020.

\bibitem{islam2020much}
M.~A. Islam, S.~Jia, and N.~D. Bruce, ``How much position information do
  convolutional neural networks encode?'' \emph{Int. Conf. Learn. Represent.},
  2020.

\bibitem{kayhan2020translation}
O.~S. Kayhan and J.~C.~v. Gemert, ``On translation invariance in cnns:
  Convolutional layers can exploit absolute spatial location,'' in \emph{IEEE
  Conf. Comput. Vis. Pattern Recog.}, 2020.

\bibitem{vaswani2017attention}
A.~Vaswani, N.~Shazeer, N.~Parmar, J.~Uszkoreit, L.~Jones, A.~N. Gomez,
  {\L}.~Kaiser, and I.~Polosukhin, ``Attention is all you need,'' in \emph{Adv.
  Neural Inform. Process. Syst.}, 2017.

\bibitem{wang2021multi}
Y.~Wang, T.~Lu, Y.~Zhang, and Y.~Wu, ``Multi-scale self-calibrated network for
  image light source transfer,'' in \emph{IEEE Conf. Comput. Vis. Pattern
  Recog. Worksh.}, 2021.

\bibitem{zhao2016loss}
H.~Zhao, O.~Gallo, I.~Frosio, and J.~Kautz, ``Loss functions for image
  restoration with neural networks,'' \emph{IEEE Trans. Comput. Imaging}, 2016.

\bibitem{hu2020sa}
Z.~Hu, X.~Huang, Y.~Li, and Q.~Wang, ``Sa-ae for any-to-any relighting,'' in
  \emph{Eur. Conf. Comput. Vis.}, 2020.

\bibitem{liu2015faceattributes}
Z.~Liu, P.~Luo, X.~Wang, and X.~Tang, ``Deep learning face attributes in the
  wild,'' in \emph{Int. Conf. Comput. Vis.}, 2015.

\bibitem{glorot2010understanding}
X.~Glorot and Y.~Bengio, ``Understanding the difficulty of training deep
  feedforward neural networks,'' in \emph{Int. Conf. Artif. Intell. Stat.},
  2010.

\bibitem{Adam15}
D.~P. Kingma and J.~Ba, ``Adam: A method for stochastic optimization,'' in
  \emph{Int. Conf. Learn. Represent.}, 2015.

\bibitem{wang2020deep}
L.-W. Wang, W.-C. Siu, Z.-S. Liu, C.-T. Li, and D.~P. Lun, ``Deep relighting
  networks for image light source manipulation,'' \emph{IEEE Conf. Comput. Vis.
  Pattern Recog. Worksh.}, 2020.

\bibitem{li2020multi}
A.~Li, Z.~Yuan, Y.~Ling, W.~Chi, C.~Zhang \emph{et~al.}, ``A multi-scale guided
  cascade hourglass network for depth completion,'' in \emph{Winter Conf. App.
  Comput. Vis.}, 2020.

\bibitem{wang2004image}
Z.~Wang, A.~C. Bovik, H.~R. Sheikh, and E.~P. Simoncelli, ``Image quality
  assessment: from error visibility to structural similarity,'' \emph{IEEE
  Trans. Image Process.}, 2004.

\bibitem{zhang2018unreasonable}
R.~Zhang, P.~Isola, A.~A. Efros, E.~Shechtman, and O.~Wang, ``The unreasonable
  effectiveness of deep features as a perceptual metric,'' in \emph{IEEE Conf.
  Comput. Vis. Pattern Recog.}, 2018.

\bibitem{jinjin2020pipal}
G.~Jinjin, C.~Haoming, C.~Haoyu, Y.~Xiaoxing, J.~S. Ren, and D.~Chao, ``Pipal:
  a large-scale image quality assessment dataset for perceptual image
  restoration,'' in \emph{Eur. Conf. Comput. Vis.}, 2020.

\bibitem{ahn2018fast}
N.~Ahn, B.~Kang, and K.-A. Sohn, ``Fast, accurate, and lightweight
  super-resolution with cascading residual network,'' in \emph{Eur. Conf.
  Comput. Vis.}, 2018.

\end{thebibliography}
\ifCLASSOPTIONcaptionsoff
  \newpage
\fi

\end{document}